\pdfoutput=1

\documentclass{article} % For LaTeX2e

\usepackage{hyperref}

\usepackage{url}
\usepackage{bbm}

\usepackage[utf8]{inputenc} % allow utf-8 input
\usepackage[T1]{fontenc}    % use 8-bit T1 fonts
\usepackage{url}            % simple URL typesetting
\usepackage{booktabs}       % professional-quality tables
\usepackage{amsfonts}       % blackboard math symbols
\usepackage{nicefrac}       % compact symbols for 1/2, etc.
\usepackage{microtype}      % microtypography
\usepackage{xcolor}         % colors

\usepackage{graphicx}
\usepackage{subfigure}

\usepackage{bm}
\usepackage{caption,stackengine}
\usepackage{adjustbox}
\usepackage{comment}
\usepackage{multirow}
\usepackage{wrapfig}

\usepackage{amsmath}
\usepackage{amssymb}
\usepackage{mathtools}
\usepackage{amsthm}

\usepackage[accepted]{icml2024}

\usepackage{paralist, tabularx}

\usepackage{pifont}% http://ctan.org/pkg/pifont
\newcommand{\cmark}{\ding{51}}%
\newcommand{\xmark}{\ding{55}}%

\newcommand\csm[1]{{\textcolor{black}{#1}}}
\newcommand\csmcm[1]{{\textcolor{black}{#1}}}

\usepackage[capitalize,noabbrev]{cleveref}

\theoremstyle{plain}

\theoremstyle{definition}

\theoremstyle{remark}

\usepackage[textsize=tiny]{todonotes}

\icmltitlerunning{Regularizing with Pseudo-Negatives for Continual Self-Supervised Learning}

\begin{document}

\twocolumn[
\icmltitle{Regularizing with Pseudo-Negatives for Continual Self-Supervised Learning}




\begin{icmlauthorlist}
\icmlauthor{Sungmin Cha}{nyu}
\icmlauthor{Kyunghyun Cho}{nyu,gen}
\icmlauthor{Taesup Moon}{snu}
\end{icmlauthorlist}

\icmlaffiliation{nyu}{New York University}
\icmlaffiliation{gen}{Genentech}
\icmlaffiliation{snu}{ASRI / INMC / IPAI / AIIS, Seoul National University}

\icmlcorrespondingauthor{Taesup Moon}{tsmoon@snu.ac.kr}

\icmlkeywords{Machine Learning, ICML}

\vskip 0.3in
]



\printAffiliationsAndNotice{}  % leave blank if no need to mention equal contribution

\begin{abstract}

% We introduce a novel and general loss function, called Augmented Negatives (AugNeg), for effective continual self-supervised learning (CSSL). We first argue that the conventional loss form of continual learning which consists of single task-specific loss (for plasticity) and a regularizer (for stability) may not be ideal for contrastive loss based CSSL that focus on representation learning. Our reasoning is that, in contrastive learning based methods, the task-specific loss would suffer from decreasing diversity of negative samples and the regularizer may hinder learning new distinctive representations. To that end, we propose AugNeg that consists of two losses with symmetric dependence on current and past models' negative representations. We argue our model can naturally find good trade-off between the plasticity and stability without any explicit hyperparameter tuning. 
% Furthermore, we present that the idea of utilizing augmented negatives can be applied to CSSL with non-contrastive learning by adding an additional regularization term.
% We validate the effectiveness of our approach through extensive experiments, demonstrating that applying the AugNeg loss achieves superior performance compared to other state-of-the-art methods, in both contrastive and non-contrastive learning algorithms.

\csmcm{We introduce a novel Pseudo-Negative Regularization (PNR) framework for effective continual self-supervised learning (CSSL).
 Our PNR leverages pseudo-negatives obtained through model-based augmentation in a way that newly learned representations may not contradict what has been learned in the past. Specifically, for the InfoNCE-based contrastive learning methods, we define symmetric pseudo-negatives obtained from current and previous models and use them in both main and regularization loss terms. 
 Furthermore, we extend this idea to non-contrastive learning methods which do not inherently rely on negatives. 
 For these methods, a pseudo-negative is defined as the output from the previous model for a differently augmented version of the anchor sample and is asymmetrically applied to the regularization term. 
 Extensive experimental results demonstrate that our PNR framework achieves state-of-the-art performance in representation learning during CSSL by effectively balancing the trade-off between plasticity and stability. }
 
\end{abstract}

\section{Introduction}

Self-Supervised Learning (SSL) has recently emerged as a cost-efficient approach for training neural networks, eliminating the need for laborious data labelling~\citep{(ssl_survey)gui2023survey}. 
Specifically, the representations learned by recent SSL methods (\textit{e.g.,} MoCo~\citep{(moco)he2020momentum}, SimCLR~\citep{(simclr)chen2020simple},  BarlowTwins~\citep{(barlow_twins)zbontar2021barlow}, BYOL~\citep{(byol)grill2020bootstrap}, and {VICReg~\citep{(vicreg)bardes2022vicreg}})
are shown to have excellent quality, comparable to those learned from supervised learning. Despite such success, huge memory and computational complexities are the apparent bottlenecks for easily maintaining and updating the self-supervised learned models, {since they typically require large-scale unsupervised data, large mini-batch sizes, and numerous gradient update steps for training}. 
 
To that end, Continual Self-Supervised Learning (CSSL), in which the aim is to learn progressively improved representations from a sequence of unsupervised data, can be an efficient alternative to the high-cost, jointly trained self-supervised learning. With such motivation, several recent studies \citep{(continuity)madaan2022representational, (stream)hu2022how, (cassle)fini2022self} have considered the CSSL using various SSL methods and showed their effectiveness in maintaining representation continuity. Despite the positive results, we note that the core idea for those methods is mainly borrowed from the large body of continual learning research for supervised learning \citep{(cl_survey1)parisi2019continual, (cl_survey2)delange2021continual, (cl_survey3)wang2023comprehensive}. Namely, a typical supervised continual learning method can be generally described as employing a single-task loss term for the new task (\textit{e.g.,} cross-entropy or supervised contrastive loss~\citep{(sup_contrastive)khosla2020supervised}) together with a certain type of 
regularization (\textit{e.g.,} distillation-based~\citep{(lwf)li2017learning,(podnet)douillard2020podnet, (afc)kang2022class, (foster)wang2022foster, (co2l)cha2021co2l} or norm-based~\citep{(ewc)kirkpatrick2017overcoming, (MAS)aljundi2018memory, (AGS-CL)jung2020adaptive, (UCL)ahn2019uncertainty,(cpr)cha2021cpr} or replay-sample based terms \citep{(bic)wu2019large, (icarl)rebuffi2017icarl}) to prevent forgetting; the recent state-of-the-art CSSL methods simply follow that approach with \textit{self-supervised} loss terms \csmcm{(\textit{e.g.}, CaSSLe~\cite{(cassle)fini2022self})}.

In this regard, we raise an issue on the current CSSL approach; the efficacy of simply incorporating a regularization term into the existing self-supervised loss for achieving successful CSSL remains uncertain. Namely, typical regularization terms are essentially designed to maintain the representations of the previous model, but they may hinder the capability of learning better representations while learning from the new task \csmcm{(\textit{i.e., plasticity)}} ~\citep{(repeval)cha2023objective, kim2023stability}. 

\csm{To address these limitations, we propose a novel method called Pseudo-Negative Regularization (PNR) for CSSL, which utilizes \textit{pseudo-negatives} for each anchor of a given input, obtained by model-based augmentation. 
In the CSSL using various SSL methods, PNR defines different pseudo-negatives tailored for each contrastive and non-contrastive learning method, respectively.
Firstly, we consider the case of using InfoNCE-type contrastive loss~\cite{(infonce)oord2018representation}, such as SimCLR and MoCo, which explicitly leverages negative samples. 
\csmcm{We propose novel loss functions by modifying the ordinary InfoNCE loss and contrastive distillation~\cite{(cont_distill1)tian2019contrastive}, ensuring that the former considers negatives from the previous model and the latter includes negatives from the current model as the \csmcm{pseudo-negatives}.}
% In other words, the proposed loss functions consider the same set of the negatives from both the current and previous models at the same time. 
As a result, these loss functions ensure that newly acquired representations do not overlap with previously learned ones, enhancing \csmcm{\textit{plasticity}}. Moreover, they enable effective distillation of prior knowledge into the current model without interfering with the representations already learned by the current model, thereby improving overall \csmcm{\textit{stability}}. 
Second, we extend the idea of using pseudo-negatives to CSSL using non-contrastive learning methods like BYOL, BarlowTwins, and VICReg, which \csmcm{do not} explicitly employ negative samples in their original implementations. 
\csmcm{For this, we reevaluate the relationship between the anchor of the current model and the negatives from the previous model observed in the contrastive distillation}
Building on this relationship, we propose a novel regularization that defines the \csmcm{pseudo-negative} for the anchor from the current model as the output feature of the same image with different augmentations from the previous model. Our final loss function aims to minimize their similarity by incorporating this regularization alongside the existing distillation term for CSSL.
Finally, through extensive experiments, our proposed method not only achieves state-of-the-art performance in CSSL scenarios and downstream tasks but also shows both better \textit{stability} and \textit{plasticity}.}

% In summary, our core contributions are summarized as below:

% \begin{compactitem}
% \item \csm{We propose Pseudo-Negative Regularization (PNR) exclusively for CSSL, specifically tailored for each contrastive and non-contrastive learning SSL method.}
% \item \csm{To employ the PNR, we introduce novel loss functions based on InfoNCE and contrastive distillation for contrastive learning, and a novel regularization for non-contrastive learning.}
% \item \csm{We experimentally confirm that the proposed loss functions using pseudo negative achieves state-of-the-art performance acoss various CSSL scenarios and downstream tasks.}
% \end{compactitem}
% \vspace{-0.1in}
\section{Related Work}

\noindent\textbf{Self-supervised representation learning} \ \
There have been several recent variations for Self-Supervised Learning (SSL)~\citep{(pretext1)alexey2016discriminative, (pretext2)doersch2015unsupervised, (denoising_autoencoder)vincent2010stacked, (colrization)zhang2016colorful, (cont_init)hadsell2006dimensionality, (simclr)chen2020simple, (moco)he2020momentum}.
Among those, 
{\textit{contrastive loss-based} methods have emerged as one of the leading approaches to learn discriminative representations~\citep{(cont_init)hadsell2006dimensionality, (infonce)oord2018representation}, in which the representations are learned by pulling the positive pairs together and pushing the negative samples apart.
Several efficient contrastive learning methods, like MoCo~\citep{(moco)he2020momentum,(mocov2)chen2020improved}, SimCLR~\citep{(simclr)chen2020simple}, have been proposed build on the InfoNCE loss~\citep{(infonce)oord2018representation}.}
Additionally, \textit{non-contrastive learning} methods, such as Barlow Twins~\citep{(barlow_twins)zbontar2021barlow}, BYOL~\citep{(byol)grill2020bootstrap} and {VICReg~\citep{(vicreg)bardes2022vicreg}, have been demonstrated to yield high-quality learned representations without \csmcm{using negative samples}.}

\noindent\textbf{Continual learning}  \ \
Continual learning (CL) is the process of acquiring new knowledge while retaining previously learned knowledge \citep{(cl_survey1)parisi2019continual, (cil_survey)masana2020class} from a sequence of tasks. 
To balance the trade-off between \textit{plasticity}, the ability to learn new tasks well, and \textit{stability}, the ability to retain  knowledge of previous tasks~\citep{(tradeoff)mermillod2013stability}, the {supervised} CL research has been proposed in several categories.
For more details, one can refer to ~\citep{(cl_survey3)wang2023comprehensive, (cl_survey2)delange2021continual}. 

\noindent\textbf{Continual Self-Supervised Learning}  \ \
Recently, there has been a growing interest in Continual Self-Supervised Learning (CSSL), as evidenced by several related researches \citep{(unsup_representation_cl)rao2019continual, (continuity)madaan2022representational, (stream)hu2022how, (cassle)fini2022self}. 
While all of them explore the possibility of using unsupervised datasets for CL, they differ in their perspectives.
\cite{(unsup_representation_cl)rao2019continual} is the first to introduce the concept of unsupervised continual learning and proposed a novel approach to learning class-discriminative representations without any knowledge of task identity.
\cite{(continuity)madaan2022representational} \csmcm{proposes a novel data augmentation method for CSSL} and first demonstrates that CSSL can outperform supervised CL algorithms in the task-incremental learning scenario.
Another study~\citep{(stream)hu2022how} focuses on the benefits of CSSL in large-scale datasets (\textit{e.g.}, ImageNet), demonstrating that a competitive pre-trained model can be obtained through CSSL.
\csmcm{The first significant regularization for CSSL was proposed by CaSSLe~\cite{(cassle)fini2022self}.
They devised a novel regularization that helps to overcome catastrophic forgetting in CSSL, achieving state-of-the-art performance in various scenarios without using exemplar memory.
After that, several papers have been published but they consider settings that are different from CaSSLe. C$^2$ASR~\citep{cheng2023contrastive} considers to use the exemplar memory and introduces both a novel loss function and exemplar sampling strategy.
\cite{yu2024evolve, tang2024kaizen, gomez2024plasticity} are tailored for semi-supervised learning scenarios and demonstrate superior performance in such cases. Additionally, \cite{yu2024evolve, gomez2024plasticity} are dynamic architecture-based algorithms where the model expands as the number of tasks grows.}

\csmcm{In this paper, we offer a few distinctive contributions compared to above mentioned related works. 
First, we identify shortcomings in the conventional regularization-based loss formulation for CSSL, such as CaSSLe. 
Second, we introduce a novel concept of pseudo-negatives and propose a new loss function that incorporates this concept, applicable to both contrastive and non-contrastive learning-based CSSL.}

% Second, we introduce two novel forms of a loss function, that leverage pseudo negative, applied to both contrastive and non-contrastive learning SSL methods, respectively. It's important to note that these proposed loss functions are specifically designed for CSSL, setting them apart from CaSSLe, which proposes the application of existing SSL methods for distillation, and C$^2$ASR, which utilizes exemplar memory for successful CSSL.

\section{Problem Setting}\label{sec:setting}
\noindent\textbf{Notations and preliminaries.}  \ \
We evaluate the quality of CSSL methods using the setting and data as in \cite{(cassle)fini2022self}. Namely, let $t$ be the task index, where $t \in \{1, \dots, T\}$, and $T$ represent the maximum number of tasks. The input data and their corresponding true labels given at the $t$-th task are denoted by $ \bm x \in \mathcal{X}_t$ and $y \in \mathcal{Y}_t$, respectively\footnote{For concreteness, we explicitly work with image data in this paper, but we note that our method is general and not confined to image modality.}. 
Let $\mathcal{D}$ is the entire dataset. We assume each training dataset for task $t$ comprises 
$M$ supervised pairs, denoted as $\mathcal{D}_t = \{ (\bm x_i, y_i)\}^{M}_{i=1}$, in which each pair is considered to be sampled from a joint distribution $p(\mathcal{X}_t, \mathcal{Y}_t)$.
Note in the case of continual supervised learning (CSL)~\citep{(cl_survey2)delange2021continual, (cil_survey)masana2020class}, both inputs and the labels are used, whereas in CSSL~\citep{(cassle)fini2022self,(continuity)madaan2022representational}, only input data are utilized for training, while the true labels are used only for the evaluation of the learned representations, such as linear probing or $k$-NN evaluation~\citep{(cassle)fini2022self, (repeval)cha2023objective}. 
Let \csmcm{$m_{\bm \psi_{t}} \circ h_{\bm \theta_t} $  is the model consisting of the representation encoder (with parameter $\bm \theta_t$) and an MLP layer (with parameter $\bm \psi_{t}$) learned after task $t$.}
To evaluate the quality of $h_{\bm \theta_{t}}$ via linear probing, we consider a classifier $f_{\bm \Theta_{t}}= o_{\bm \phi_{t}}\circ h_{\bm \theta_{t}}$, in which $\bm \Theta_t=(\bm \theta_t,\bm \phi_t)$ and $o_{\bm \phi_{t}}$ is the linear output layer (with parameter $\bm \phi_t$) on top of $h_{\bm \theta_{t}}$. Then, only $o_{\bm \phi_{t}}$ is supervised trained (with frozen $h_{\bm \theta_{t}}$) using all the training dataset $\mathcal{D}_{1:t}$, including the labels, and the accuracy of resulting $f_{\bm \Theta_{t}}$ becomes the proxy for the representation quality.

\noindent\textbf{Class-/Data-/Domain-incremental learning.}  \ \
We consider the three scenarios of continual learning as outlined in \cite{(CL_scenarios)van2019three, (cl_survey3)wang2023comprehensive, (cassle)fini2022self}. We use $k$ and $j$ to denote arbitrary task numbers, where $k,j\in \{1,\ldots,T\}$ and $k \neq j$. 
The first category is the \textit{class-incremental learning} (Class-IL), in which the $t$-th task's dataset consists of a unique set of classes for the input data, namely, $p(\mathcal{X}_k)\neq p(\mathcal{X}_j)$ and $\mathcal{Y}_k \cap \mathcal{Y}_j = \varnothing$.
The second category is \textit{domain-incremental learning} (Domain-IL), in which each dataset $\mathcal{D}_t$ has the same set of true labels but with different distribution on $\mathcal{X}_t$, denoted as $p(\mathcal{X}_k)\neq p(\mathcal{X}_j)$ but $\mathcal{Y}_k = \mathcal{Y}_j$.
In other words, each dataset in Domain-IL contains input images sampled from a different domain, but the corresponding set of true labels is the same as for other tasks.
Finally, we consider \textit{data-incremental learning} (Data-IL), in which a set of input images $\mathcal{X}_t$ is sampled from a single distribution, $p(\mathcal{X}_k) = p(\mathcal{X}_j)$, but $\mathcal{Y}_k = \mathcal{Y}_j$. 
To implement the Data-IL scenario in our experiments, we shuffle the entire dataset (such as ImageNet-100) and divide it into $T$ disjoint datasets.

\begin{figure}[t]
% \vspace{-.05in}
\centering 
{\includegraphics[width=0.95\linewidth]{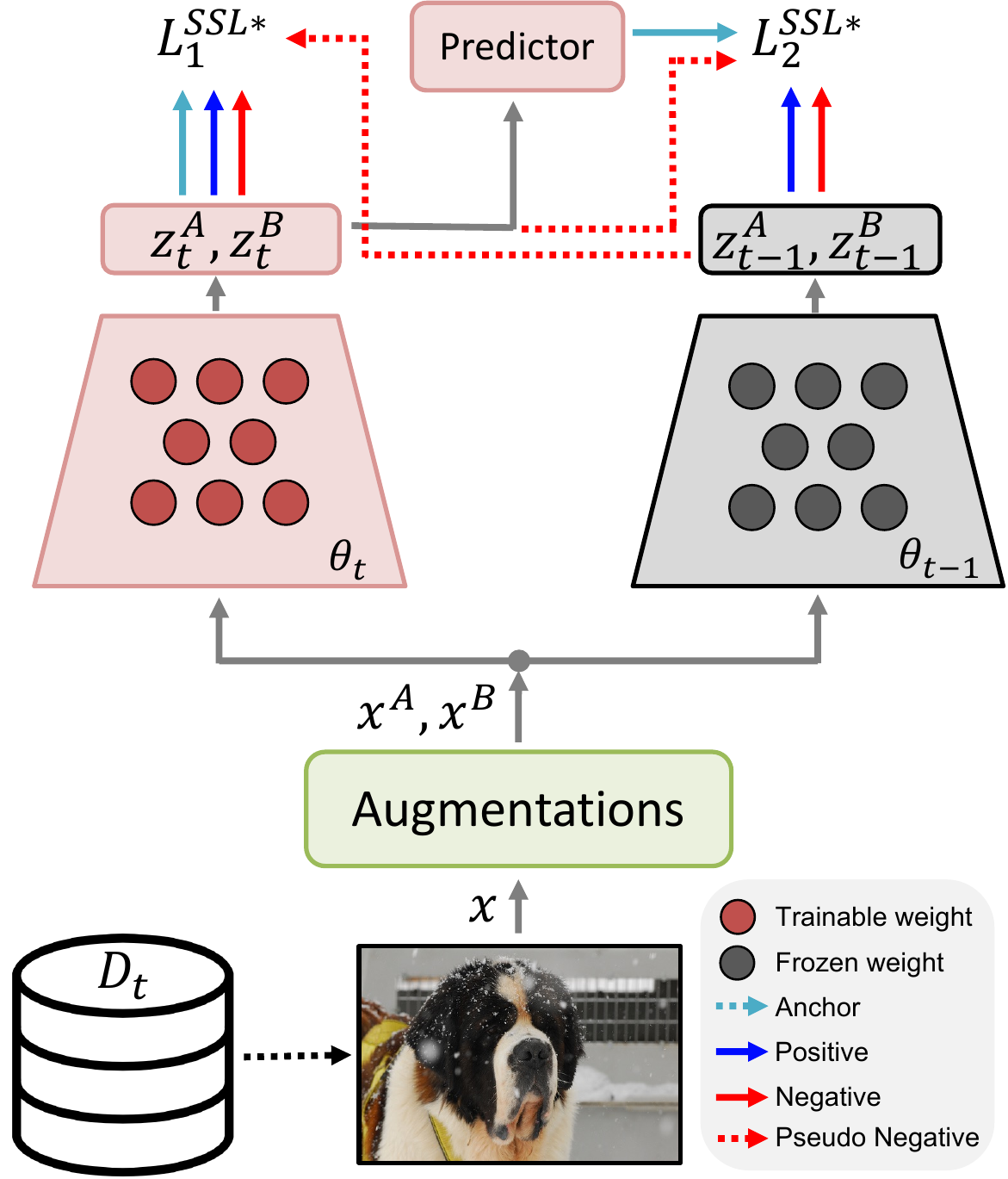}}
\vspace{-.05in}
\caption{The overview of using pseudo-negatives in CSSL with contrastive learning. Note that red dashed arrows denote the incorporation of the proposed \csmcm{pseudo-negatives}, which are output features from distinct models,  in each loss function.}
\vspace{-.15in}
\label{figure:motivation}
 \end{figure}

\section{Pseudo-Negative Regularization (PNR)}

\subsection{Motivation}\label{sec:motivation}
Several studies have been conducted with the aim of progressively improving the quality of representations learned by the encoder ($h_{\bm \theta_{t}}$) in CSSL. 
While it has been noted that simple fine-tuning using $D_t$ results in less severe forgetting compared to  supervised continual learning~\cite{(continuity)madaan2022representational, davari2022probing}, CaSSLe~\cite{(cassle)fini2022self} has achieved even more successful CSSL.
\csmcm{It introduces a novel regularization method based on existing SSL methods, using the output features of both the encoder at time $t$ and the encoder at time $t-1$, to overcome catastrophic forgetting in learned representations.}
Despite CaSSLe achieving promising results in various experiments, our motivation arises from the belief that incorporating \textit{pseudo-negatives} can lead to even more successful results. Consider an augmented image obtained by applying different augmentations to the input image $\bm x$, denoted by $\bm x^A$ and $\bm x^B$. 
% The output features of the $h_{\bm \theta_t}$ and $h_{\bm \theta_{t-1}}$ encoders 
\csmcm{The output features of the $m_{\bm \psi_{t}} \circ h_{\bm \theta_t} $ and $m_{\bm \psi_{t-1}} \circ h_{\bm \theta_t-1} $} 
for these augmented images are denoted by $\bm z^A_{t}$, $\bm z^B_{t}$, $\bm {z}^A_{t-1}$, and $\bm {z}^B_{t-1}$, respectively. 
In contrast to CaSSLe, our approach involves utilizing the output features of both the models as pseudo-negatives, as illustrated in Figure \ref{figure:motivation}. For this purpose, we propose a novel CSSL loss form for task $t$, defined as follows:
% \begin{equation}
% \begin{aligned}
% \mathcal{L}_{t}^{\text{CSSL}}(\bm x^A, \bm x^B; \bm \theta_t, \bm \theta_{t-1}) &= \mathcal{L}^{\text{SSL}^*}_1(z^A, z^B, \bar{z}^A, \bar{z}^B) \\ &+  \mathcal{L}^{\text{SSL}^*}_2(g(z^A), \bar{z}^A, z^A, z^B),  \label{eqn:recon}
% \end{aligned}
% \end{equation} 
\begin{equation}
\begin{aligned}
&\mathcal{L}_{t}^{\text{CSSL}}(\{\bm x^A, \bm x^B\};\bm \theta_t, \bm \theta_{t-1})\\ 
&=
\mathcal{L}^{\text{SSL}^*}_1(\{\bm z_t^A, \bm z_t^B, \bm z_{t-1}^A, \bm z_{t-1}^B\})\\&+  \mathcal{L}^{\text{SSL}^*}_2(\{g(\bm z_t^A), \bm z_{t-1}^A, \bm z_{t-1}^B, \bm z_t^A, \bm z_t^B\}).  
\end{aligned}\label{eqn:cont_form}
\end{equation} 
\csmcm{Here, $g(\cdot)$ represents another MLP layer (referred to as the Predictor in the figure) introduced in \cite{(cassle)fini2022self}, which has the same shape as $m_{\bm \psi_{t}}$.}
The SSL loss function $\mathcal{L}_{1/2}^{\text{SSL}^{*}}$ are newly designed general loss forms incorporating all candidate pseudo-negatives: $\bm z_{t-1}^A, \bm z_{t-1}^B$ for $\mathcal{L}^{\text{SSL}^{*}}_1$ and $\bm z_t^A, \bm z_t^B$ for $\mathcal{L}^{\text{SSL}^{*}}_2$. They resemble the two loss functions utilized in CaSSLe, and we will highlight the specific difference later. Furthermore, in order to 
\csmcm{consider two different augmentations in a symmetric fashion, we employ the following average as the final loss function for our method $\frac{1}{2}(\mathcal{L}_{t}^{\text{CSSL}}(\bm x^A, \bm x^B) + \mathcal{L}_{t}^{\text{CSSL}}(\bm x^B, \bm x^A))$.}
% The overall loss for CSSL is symmetric, such as $\frac{1}{2}(\mathcal{L}_{t}^{\text{CSSL}}(\bm x^A, \bm x^B) + \mathcal{L}_{t}^{\text{CSSL}}(\bm x^B, \bm x^A))$.

\begin{figure}[t]
\vspace{-.05in}
\centering 
{\includegraphics[width=0.98\linewidth]{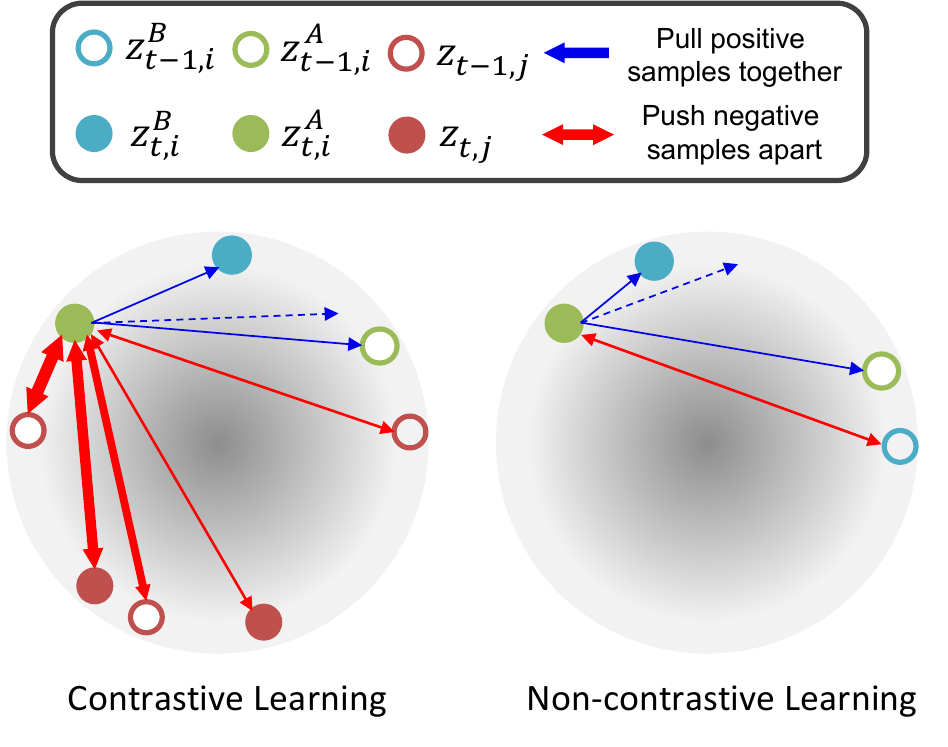}}
\vspace{-.2in}
\caption{Graphical representation of learning with our proposed loss. The blue dashed arrow indicates the direction of the gradient update during training with the proposed loss. It moves away from the negative and pseudo-negative embeddings, which correspond to current and past models, while converging towards the positive embeddings of the current and past models.}
\vspace{-.15in}
\label{figure:illustration}
 \end{figure}

As mentioned above, CaSSLe also uses two SSL loss functions, one responsible for plasticity and the other for stability, but they do not utilize any pseudo-negatives. Namely, the plasticity loss of CaSSLe (which corresponds to $\mathcal{L}^{\text{SSL}^{*}}_1$ of ours) is designed to learn new representations from the new task $t$ and only utilizes the output features from the current model $t$. Moreover, the stability loss of CaSSLe (which corresponds to $\mathcal{L}^{\text{SSL}^{*}}_2$ of ours) aims to maintain the representations learned from the past task $t-1$ and only regularizes with the output features from the previous model $t-1$.
% that the CaSSLe does not use the pseudo-negatives in both loss functions ($\mathcal{L}^{\text{SSL}^{*}}_1$ and $\mathcal{L}^{\text{SSL}^{*}}_2$) and leverages existing SSL methods to achieves plasticity and stability in CSSL.
% Specifically, for the identical anchor of the model $t$, the CaSSLe's $\mathcal{L}^{\text{SSL}^{*}}_1$ aims to learn new representations (\textit{i.g.}, plasticity) from the new task by minimizing the SSL loss through the sole use of positive and negative pairs from output features from the model $t$. 
% In contrast, the CaSSLe's $\mathcal{L}^{\text{SSL}^{*}}_2$ aims to maintain learned representations (\textit{i.g.}, stability) by employing positive and negative pairs from the output features of the model $t-1$.
We argue that such a stability loss term, designed to preserve representations from the previous model, might impede the plasticity loss, that aims to learn new representations from the new task but does not consider any representations from the previous model~\citep{(repeval)cha2023objective, kim2023stability}. 
% ability to learn new representation from the new task 
% improved representations when \csmcm{learning with} new task data~\citep{(repeval)cha2023objective, kim2023stability}. Moreover, the divergent configuration of positive and negative pairs for the same anchor could lead to conflicts in the learned representations, potentially hindering the acquisition of superior representations. 
To address these issues, we propose utilizing the output features of both the model $t-1$ and $t$ as the \textit{pseudo-negatives}, that can effectively work as regularizers so that the newly learned representations may not interfere with previously learned representations.

% ensuring that each loss term learns representations from the same configuration of negatives.

In the upcoming section, we will specifically introduce novel function forms for $\mathcal{L}^{\text{SSL}^{*}}_1$ and $\mathcal{L}^{\text{SSL}^{*}}_2$ tailored to integrate pseudo-negatives into contrastive learning methods (e.g., SimCLR~\cite{(simclr)chen2020simple}, MoCo~\cite{(moco)he2020momentum}). Additionally, we will present how the idea of using pseudo-negatives can be applied to non-contrastive learning methods (e.g., BYOL~\cite{(byol)grill2020bootstrap}, VICReg~\cite{(vicreg)bardes2022vicreg}, BarlowTwins~\cite{(barlow_twins)zbontar2021barlow}) as well.

\subsection{InfoNCE-based Contrastive Learning Case}\label{sec:method_contrastive} 
Here, we propose new loss functions for CSSL using contrastive learning-based SSL methods (\textit{e.g.}. SimCLR and MoCo). First, $\mathcal{L}^{\text{SSL}^*}_1$ in (\ref{eqn:cont_form}) is defined as:
\begin{equation}
\begin{aligned}
&\mathcal{L}^{\text{SSL}^{*}}_1(\{\bm z_t^A, \bm z_t^B, \bm z_{t-1}^A, \bm z_{t-1}^B\})\\&=- {\mathrm{log}}\frac{\mathrm{exp}(\bm z^A_{t,i} \cdot \bm z^{B}_{t,i} / \tau)}{\sum_{\bm z_j\in\mathcal{N}_1(i)\cup\mathcal{PN}_1(i)}\exp (\bm z^A_{t,i}\cdot \bm z_j/\tau)},
\label{eqn:plasticity}
\end{aligned}
\end{equation} 
in which 
$\mathcal{N}_1(i)=\{\bm z_{t}^A, \bm z_{t}^B\}\backslash \{z_{t,i}^A\}$ is the set of original negatives,
% $\mathcal{N}(i)=\{\bm z_{t,k}^A\}_{k\neq i} \cup \bm z_{t,k}^B$ 
and
$\mathcal{PN}_1(i)=\{\bm z_{t-1}^A, \bm z_{t-1}^B\}\backslash \{z_{t-1,i}^A\}$ are pseudo-negatives of $\mathcal{L}^{\text{SSL}^{*}}_1$.
% $\mathcal{PN}(i)=\{\bm z_{t-1,k}^A\}_{k\neq i} \cup \bm z_{t-1,k}^B$. 
Also, $\mathcal{L}^{\text{SSL}^*}_2$ is defined as follows:
\begin{equation}
\begin{aligned}
&\mathcal{L}^{\text{SSL}^{*}}_2(\{g(\bm z_t^A), \bm z_{t-1}^A, \bm z_{t-1}^B, \bm z_t^A, \bm z_t^B\})\\&=- {\mathrm{log}}\frac{\mathrm{exp}(g(\bm z^A_{t,i}) \cdot \bm z^{A}_{t-1,i} / \tau)}{\sum_{\bm z_j\in\mathcal{N}_2(i)\cup\mathcal{PN}_2(i)}\exp (g(\bm z^A_{t,i})\cdot \bm z_j/\tau)},
\label{eqn:stability}
\end{aligned}
\end{equation} 
in which 
$\mathcal{N}_2(i)=\{\bm z_{t-1}^A, \bm z_{t-1}^B\}\backslash \{z_{t-1,i}^A\}$ is the original negatives for contrastive distillation~\cite{(cont_distill1)tian2019contrastive, (cassle)fini2022self},
% $\mathcal{N}(i)=\{\bm z_{t-1,k}^A\}_{k\neq i} \cup \bm z_{t-1,k}^B$ 
and 
$\mathcal{PN}_2(i)=\{\bm z_{t}^A, \bm z_{t}^B\}\backslash \{z_{t,i}^A\}$ is the pseudo-negatives of $\mathcal{L}^{\text{SSL}^{*}}_2$.
% $\mathcal{PN}(i)=\{\bm z_{t,k}^A\}_{k\neq i} \cup \bm z_{t,k}^B$. 
Also, $\tau$ denotes a temperature parameter.
% and $\mathrm{exp}$ stands for an exponential function. 
In the case of SimCLR, $\mathcal{N}_1(i)$, $\mathcal{PN}_1(i)$, $\mathcal{N}_2(i)$, and $\mathcal{PN}_2(i)$ consist of negatives from the current batch. When using MoCo, two queues storing positives of each loss function ($\mathcal{L}^{\text{SSL}^{*}}_1$ and $\mathcal{L}^{\text{SSL}^{*}}_2$) from previous iterations  are employed for them.

Note these two losses are quite similar in form to InfoNCE~\citep{(infonce)oord2018representation}, have has a couple of key differences. 
First, in $\mathcal{L}_{\text{1}}^{\text{SSL}^*}$, we use pseudo-negatives in $\mathcal{PN}_1(i)$ which are obtained from the previous model $h_{\bm \theta_{t-1}}$. 
This addition of negative embeddings in $\mathcal{L}_{\text{1}}^{\text{SSL}^*}$ compels the embedding of $\bm x_i$ to be repelled not only from the negative embeddings of the current model but also from those of the previous model, hence, it fosters the acquisition of more distinctive representations. 
Second, in $\mathcal{L}_{\text{2}}^{\text{SSL}^*}$, which has the similar form of contrastive distillation, we also use pseudo-negatives in $\mathcal{PN}_2(i)$. Such modification has the impact of placing additional constraints on distillation, ensuring that the representations from the past model are maintained in a way that does not contradict the representations of the current model.
% Also, we introduce the ``Predictor'' layer, as in \cite{(cassle)fini2022self}, to not hurt the plasticity by doing indirect distillation. 
Note that the denominators of the two loss functions are identical except for the $g(\cdot)$ in $\mathcal{L}_{\text{2}}^{\text{SSL}^*}$. 
Therefore, with the identical denominators, adding two losses will  result in achieving a natural trade-off between plasticity and stability for learning the representation. 

This intuition is depicted in the left figure of Figure \ref{figure:illustration}. Namely,  both losses symmetrically consider the embeddings from current and previous models, and as shown in the hypersphere, the representation of $\bm z^{A}_{t,i}$ gets attracted to $\bm z^{B}_{t,i}$ and $\bm z^{A}_{t-1,i}$ with the constraint that it is far from $\bm z^{B}_{t-1,i}$. Thus, the new representation will be distinctive from the previous model (\textit{plasticity}) and carry over the old knowledge (\textit{stability}) in a way not hurt the current model. 

The gradient analysis of the proposed loss function is detailed in Section \ref{sec:sup_gradient} of the Appendix.

\subsection{Non-Contrastive Learning Case}\label{sec:method_non_contrastive}

\csm{Non-contrastive learning methods, such as Barlow, BYOL, and VICReg, do not incorporate negative samples. Therefore, the direct application of pseudo-negatives used for contrastive learning methods is not viable for these methods. 
However, motivated by the configuration of negatives in Equation (\ref{eqn:stability}), we propose a novel regularization that considers the pseudo-negatives from a new perspective. 
To achieve this, we propose new formulations of $\mathcal{L}^{\text{SSL}^{*}}_1$ and $\mathcal{L}^{\text{SSL}^{*}}_2$, tailored for non-contrastive learning methods, as follows:}
\begin{equation}
\begin{aligned}
&\mathcal{L}^{\text{SSL}^{*}}_1(\{\bm z_t^A, \bm z_t^B, \bm z_{t-1}^A, \bm z_{t-1}^B\}) = \mathcal{L}^{\text{SSL}}(\{z_t^A, \bm z_t^B\}),
\end{aligned}\label{eqn:noncont_l2_1}
\end{equation} 
where $\mathcal{L}^{\text{SSL}}$ denotes a non-contrastive SSL loss, and
\begin{equation}
\begin{aligned}
&\mathcal{L}^{\text{SSL}^{*}}_2(\{g(\bm z_t^A), \bm z_{t-1}^A, \bm z_{t-1}^B, \bm z_t^A, \bm z_t^B\}) \\ &= \mathcal{L}^{\text{SSL}}(\{g(\bm z_t^A), \bm z_{t-1}^A\}) - \lambda *\sum_i \sum_{\bm z_j\in\mathcal{PN}_2(i)}\|g(\bm z_{t,i}^A) - \bm z_j\|^2_2,
\end{aligned}\label{eqn:noncont_l2_2}
\end{equation} 
where $\mathcal{PN}_2(i)=\{\bm z_{t-1,i}^B\}$ is the pseudo-negative, $\|\cdot\|^2_2$ represents the squared $L_2$ norm and $\lambda$ is a hyperparameter. Additionally, $\bm z_{t-1}^A$ and $\bm z_{t-1}^B$ of Equation (\ref{eqn:noncont_l2_1}), as well as $ \bm z_t^A$ and $\bm z_t^B$ of Equation (\ref{eqn:noncont_l2_2}), are not employed when using non-contrastive learning.

Note that $\mathcal{L}^{\text{SSL}}$ of Equation (\ref{eqn:noncont_l2_2}) is the CaSSLe's distillation \csmcm{for a non-contrstive learning method}, and we introduce a new regularization to incorporate the pseudo-negatives. Specifically, $\bm z_{t-1,i}^B$ is assigned as the pseudo-negative of $\bm z_{t,i}^A$ (the anchor). This assignment stems from the configuration of negatives of Equation (\ref{eqn:stability}).
For example, when implementing Equation (\ref{eqn:stability}) using SimCLR and $N$ is the mini-batch size, $\mathcal{N}(i)$ consists of $2N-1$ negatives excluding $z_{t-1,i}^A$. Consequently, for a given anchor $\bm z_{t,i}^A$, an output feature $\bm z_{t-1, i}^B$ from the same image $\bm x_i$ but subjected to different augmentation is naturally considered as a negative (this holds true when using MoCo). This leads to minimizing the similarity between $g(\bm z_{t,i}^A)$ and $\bm z_{t-1,i}^B$ for the training task $t$.
To apply this concept of negatives to CSSL using non-contrastive learning, we propose a novel regularization that maximizes the squared mean square error between $g(\bm z_{t,i}^A)$ and $\bm z_{t-1,i}^B$, ensuring their dissimilarity.

The right figure in Figure \ref{figure:illustration} illustrates   representation learning with pseudo-negative in CSSL using BYOL. The CaSSLe's representation learning relies on distinct update directions from each positive (e.g., $\bm z_{t}^A$ and $\bm z_{t-1}^A$) to achieve enhanced plasticity and stability, without taking negatives into consideration. 
However, incorporating the pseudo-negative enables the model to avoid conflicts in learning representations by considering the pseudo-negative from the model $t-1$—learning representations far from the pseudo-negative from the model $t-1$. Consequently, the model acquires more distinctive representations from the previous model (enhancing plasticity) while retaining prior knowledge (ensuring stability).
Note that Equation (\ref{eqn:noncont_l2_2}) can be applied in conjunction with various non-contrastive SSL methods. 
The implementation details \csmcm{for BYOL and VICReg} are provided in Section \ref{sec:implementation} of the Appendix.

We will refer to the overall framework of using pseudo-negatives for regularization in CSSL, applicable to both contrastive and non-contrastive learning as described above, as PNR (Pseudo-Negative Regularization).
% We will refer the overall framework of utilizing pseudo-negatives for regularization in CSSL as PNR (Pseudo-Negative Regularization). 

% Note that the Equation \ref{eqn:noncont_form} can be applied together with several non-contrastive SSL methods, such as BYOL, VICReg and BarlowTwins. 
% We believe that the above idea can similarly be applied to various SSL algorithms that do not explicitly involve negative samples through the addition of extra regularization. The implementation details for BYOL, VICReg and Barlow are introduced in the Supplementary Materials.
\section{Experiments}

\subsection{Experimental Details}
\textbf{\noindent{Baselines}} \ \
To evaluate the proposed PNR, we set CaSSLe~\citep{(cassle)fini2022self} as our primary baseline, which has shown state-of-the-art performance in CSSL.
{We select five SSL methods, SimCLR~\citep{(simclr)chen2020simple}, MoCo v2 Plus (MoCo)~\citep{(mocov2)chen2020improved}, BarlowTwins (Barlow)~\citep{(barlow_twins)zbontar2021barlow},  BYOL~\citep{(byol)grill2020bootstrap}, and VICReg~\citep{(vicreg)bardes2022vicreg},  which achieve superior performance  with the combination with CaSSLe.}
% CaSSLe was applied to four SSL techniques: SimCLR~\citep{(simclr)chen2020simple}, MoCo v2 Plus (MoCo)~\citep{(mocov2)chen2020improved}, BarlowTwins (Barlow)~\citep{(barlow_twins)zbontar2021barlow}, and BYOL~\citep{(byol)grill2020bootstrap}. We ran their original code provided by CaSSLe without any modifications to the algorithms or their default hyperparameters. Additional details on baseline training settings can be found in the Supplementary Material.
% In order to evaluate the proposed Sy-CON loss, we have selected CaSSLe~\cite{(cassle)fini2022self} as our primary baseline, which has demonstrated state-of-the-art performance in CSSL. We present the results obtained by applying CaSSLe to four self-supervised learning techniques, including SimCLR~\cite{(simclr)chen2020simple}, MoCo v2 Plus (MoCo)\cite{(mocov2)chen2020improved}, BarlowTwins (BT)\cite{(barlow_twins)zbontar2021barlow}, and BYOL~\cite{(byol)grill2020bootstrap}, which have all exhibited superior performance with CaSSLe across various CSSL scenarios. The results are obtained by directly running their code proposed by CaSSLe, and no modifications were made to any of the algorithms or their default hyperparameters, except for the mini-batch size, which is set to the same value as Sy-CON, taking into account the hardware environment (see the implementation details section for more information). Additional details regarding the settings used to train the baseline are provided in the Supplementary Material.

\textbf{\noindent{Implementation details}} \ \
We implement our PNR based on the code provided by CaSSLe.
We conduct experiments on four datasets: CIFAR-100~\citep{(cifar)krizhevsky2009learning}, ImageNet-100~\citep{(imagenet)deng2009imagenet},  DomainNet~\citep{(domainnnet)peng2019moment}, and ImageNet-1k~\citep{(imagenet)deng2009imagenet} following the training and evaluation process outlined in \cite{(cassle)fini2022self}. For CIFAR-100 and ImageNet-100, we perform class- and data-incremental learning (Class- and Data-IL) for 5 and 10 tasks (denoted as 5T and 10T), respectively. For Domain-incremental learning (Domain-IL), we use DomainNet~\cite{(domainnnet)peng2019moment}, consisting of six disjoint datasets from different six source domains. 
Following experiments conducted by CaSSLe, we perform Domain-IL in the task order of "Real $\rightarrow$ QuickDraw $\rightarrow$ Painting $\rightarrow$ Sketch $\rightarrow$ InfoGraph $\rightarrow$ Clipart". Next, we report the average top-1 accuracy achieved by training a linear classifier separately for each domain, employing a frozen feature extractor (domain-aware evaluation).
% Due to time constraints, we only conduct experiments on Class-IL for 5T in the case of ImageNet-1000.
The ResNet-18~\citep{(resnet)he2016deep} model implemented in PyTorch is used except for ImageNet-1k where the ResNet-50 is employed. 
% {The proposed PNR is implemented based on the code provided by CaSSLe~\citep{(cassle)fini2022self}. 
% We conducted all experiments of both CaSSLe and PNR in the unified environment for fair comparison}. 
For all experiments except for ImageNet-1k, we perform each experiment using three random seeds and report the average performance across these trials.
Further experimental details are available in Section \ref{sec:details} of the Appendix.

\textbf{\noindent{Evaluation metrics}} \ \
{To gauge the quality of representations learned in CSSL, we conduct linear evaluation by training only the output layer on the given dataset while maintaining the encoder $h_{\bm \theta_t}$ as a fixed component, following \citep{(cassle)fini2022self, (repeval)cha2023objective}}. 
The average accuracy after learning the task $t$ is denoted as {A$_{t}=\dfrac{1}{t}\sum_{i=1}^{T}a_{i,t}$}, where {$a_{i,j}$} stands for the linear evaluation top-1 accuracy of the encoder on the dataset of task $i$ after the end of learning task $j$.
Furthermore, we employ measures of stability ($S$) and plasticity ($P$), and a comprehensive explanation of these terms is provided in Section \ref{sec:measure} of the Appendix.

\begin{figure}[h]
\vspace{-.1in}
\centering
\includegraphics[width=0.9\linewidth]{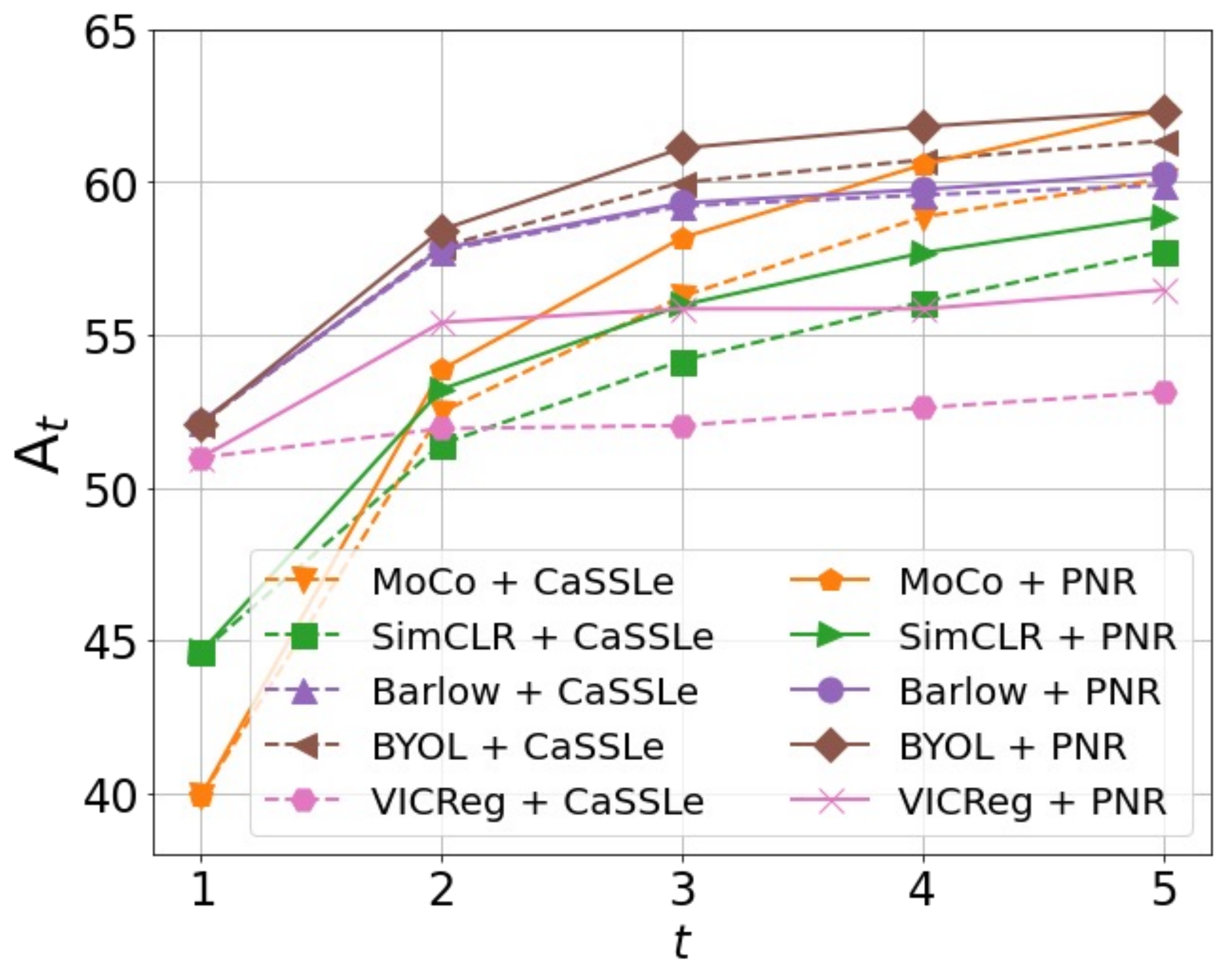}
\vspace*{-0.1in}
\caption{{Experimental results of applying PNR to SSL methods}. Note that "+CaSSLe" and "+PNR" indicate the results of applying CaSSLe and PNR to each SSL method, respectively.}\label{fig:applying}
\vspace{-.1in}
 \end{figure}
 
\subsection{Experiments with SSL Methods}

To assess the effectiveness of PNR when applied to various SSL methods, we conduct experiments by incorporating the PNR with MoCo, SimCLR, Barlow, BTOL, and VICReg, in the Class-IL (5T) scenario using CIFAR-100.
Figure \ref{fig:applying} illustrates that both the proposed PNR and CaSSLe are successfully combined with each SSL method, demonstrating progressively enhanced the quality of representations at each task.
However, PNR demonstrates more effective integration with MoCo, SimCLR, BYOL, {and VICReg}, surpassing the performance of CaSSLe.

% Firstly, when both CaSSLe and PNR are combined with each SSL method, they progressively enhance the quality of representations as each task $t$ is learned. Secondly, PNR demonstrates more effective integration with MoCo, SimCLR, BYOL, {and VICReg}, surpassing the performance of CaSSLe, which is the previous state-of-the-art method.

\begin{table}[h]
\vspace{-.1in}
\caption{{Experimental results of applying PNR to SSL methods in Class-IL (5T) with CIFAR-100. The * symbol indicates results from the CaSSLe paper, while the others are reproduced performance. The values in parentheses indicate the standard deviation and the \textbf{bolded} result represents the best performance.}}
% \vspace{-.15in}
\centering
\smallskip\noindent
% \vspace{-.1in}
\resizebox{.98\linewidth}{!}{
\begin{tabular}{c||c|c|c|c|c}
\hline
A$_{5}$        & MoCo                                                         & SimCLR                                                          & Barlow                                                          & BYOL & {VICReg}                                                            \\ \hline \hline
Joint          & \begin{tabular}[c]{@{}c@{}}66.90\\ (0.11)\end{tabular}          & \begin{tabular}[c]{@{}c@{}}63.78\\ (0.22)\end{tabular}          & \begin{tabular}[c]{@{}c@{}}68.99\\ (0.21)\end{tabular}          & \begin{tabular}[c]{@{}c@{}}69.36\\ (0.28)\end{tabular}    & \begin{tabular}[c]{@{}c@{}}68.01\\ (0.36)\end{tabular}       \\ \hline \hline
FT             & \begin{tabular}[c]{@{}c@{}}51.95\\ (0.26)\end{tabular}           & \begin{tabular}[c]{@{}c@{}}48.97\\ (0.74)\end{tabular}          & \begin{tabular}[c]{@{}c@{}}55.81\\ (0.57)\end{tabular}          & \begin{tabular}[c]{@{}c@{}}52.43\\ (0.62)\end{tabular}   & \begin{tabular}[c]{@{}c@{}}52.43\\ (0.62)\end{tabular}         \\ \hline
EWC*           & -                                                               & 53.60                                                           & 56.70                                                           & 56.40                                                       & -                                                           \\ \hline
DER*           & -                                                               & 50.70                                                           & 55.30                                                           & 54.80                                                         & -                                                           \\ \hline
LUMP*          & -                                                               & 52.30                                                           & 57.80                                                           & 56.40                                                         & -                                                           \\ \hline
Less-Forget*   & -                                                               & 52.50                                                           & 56.40                                                           & 58.60                                                        & -                                                           \\ \hline  
+CaSSLe         & \begin{tabular}[c]{@{}c@{}}60.11\\ (0.30)\end{tabular}          & \begin{tabular}[c]{@{}c@{}}57.73\\ (1.07)\end{tabular}          & \begin{tabular}[c]{@{}c@{}}60.10\\ (0.38)\end{tabular}          & \begin{tabular}[c]{@{}c@{}}61.36\\ (1.38)\end{tabular}  & \begin{tabular}[c]{@{}c@{}}53.13\\ (0.64)\end{tabular}          \\ \hline \hline
{+PNR} & {\begin{tabular}[c]{@{}c@{}}\textbf{62.36}\\ (0.29)\end{tabular}} & {\begin{tabular}[c]{@{}c@{}}\textbf{58.87}\\ (0.16)\end{tabular}} & {\begin{tabular}[c]{@{}c@{}}\textbf{60.28}\\ (0.36)\end{tabular}} & {\begin{tabular}[c]{@{}c@{}}\textbf{63.19}\\ (0.39)\end{tabular}} & {\begin{tabular}[c]{@{}c@{}}\textbf{56.47}\\ (0.94)\end{tabular}} \\ \hline
\end{tabular}\label{fig:baseline_cifar100}
}
\vspace{-.1in}
\end{table}

Table \ref{fig:baseline_cifar100} presents the numerical results of $A_{5}$ in the same scenario. In this table, "Joint" corresponds to the experimental results in the Joint SSL scenario (upper bound) and "FT" represents the results achieved through fine-tuning with the SSL method alone. 
The table reveals that PNR consistently outperforms CaSSLe, showcasing a maximum improvement of approximately 2-3\%. 
Note that the extra performance gain achieved by PNR is particularly noteworthy, especially considering that CaSSLe's performance with MoCo, SimCLR, and BYOL is already close to that of the Joint.
% Especially notable is the additional performance improvement offered by PNR, particularly when taking into account the superior performance of CaSSLe with MoCo and BYOL, which is already approaching that of the Joint. 
% We contend that the extra performance improvement achieved by PNR is a noteworthy outcome.

% {Finally, Table \ref{table:forgetting_forward} presents stability ($S$) and plasticity ($P$) metrics. For MoCo, SimCLR, BYOL, and {VICReg}, the results illustrate PNR's superiority, attributed to a significant boost in \textit{plasticity} while minimizing increase of \textit{stability} compared to CaSSLe. This underscores the crucial role of employing the proposed model-based augmentation of negatives in CSSL.

% \begin{table}[h]
% \vspace{-.2in}
% \caption{Experimental results on stability and plasticity.}
% % \vspace{-.1in}
% \label{table:forgetting_forward}
% \centering
% \smallskip\noindent
% \resizebox{.98\linewidth}{!}{
% \begin{tabular}{|c||c|c|c|c|}
% \hline
% {$S(\downarrow)$ / $P(\uparrow)$}   & MoCo        & SimCLR      & BYOL      & {VICReg}         \\ \hline \hline
% CaSSLe & 0.11 / 9.99 & 0.33 / 4.08 & 0.09 / 11.94 & 0.39 / 6.78 \\ \hline
%  PNR   & 0.06 / 11.8 & 0.48 / 5.36 & 0.20 / 13.84 & 0.65 / 12.59 \\ \hline
% \end{tabular}
% }
% \vspace{-.1in}
% \end{table}

% Based on the results of these previous experiments, we will more focus to apply and evaluate  PNR(MoCo) and  PNR(BYOL), which achieves superior performance, in further experiments on various CSSL scenarios.}

\subsection{Experiments with Diverse CSSL Scenarios}
% \vspace{-.1in}
\begin{table}[t]
\caption{{The experimental results of Class-IL. All results are reproduced performance. The values in parentheses indicate the standard deviation and the \textbf{bolded} result represents the best performance.}}
% \vspace{-.15in}
\centering
\smallskip\noindent
% \vspace{-.1in}
\resizebox{.98\linewidth}{!}{
\begin{tabular}{|cc||ccc|}
\hline
\multicolumn{2}{|c||}{}                                         & \multicolumn{3}{|c|}{Class-IL}                                                                                                               \\ \cline{3-5} 
\multicolumn{2}{|c||}{}                                         & \multicolumn{1}{|c|}{CIFAR-100}                           & \multicolumn{2}{c|}{ImageNet-100}                                                \\ \cline{3-5} 
\multicolumn{2}{|c||}{\multirow{-3}{*}{A$_{T}$}}                & \multicolumn{1}{c|}{10T}                                 & \multicolumn{1}{c|}{5T}                                  & 10T                   \\ \hline \hline
\multicolumn{1}{|c|}{}                         & Joint         & \multicolumn{1}{c|}{66.90 (0.11)}                        & \multicolumn{2}{c|}{76.67 (0.56)}                                                \\ \cline{2-5} 
\multicolumn{1}{|c|}{}                         & FT            & \multicolumn{1}{c|}{34.11 (0.90)}                        & \multicolumn{1}{c|}{57.87 (0.49)}                        & 47.48 (0.42)          \\ \cline{2-5} 
\multicolumn{1}{|c|}{}                         & +CaSSLe       & \multicolumn{1}{c|}{53.58 (0.41)}                        & \multicolumn{1}{c|}{63.49 (0.44)}                        & 52.71 (0.47)          \\ \cline{2-5} 
\multicolumn{1}{|c|}{\multirow{-4}{*}{MoCo}}   & {+PNR} & \multicolumn{1}{c|}{{\textbf{56.62} (0.31)}}               & \multicolumn{1}{c|}{{\textbf{67.85} (0.44)}}               & {\textbf{60.75} (0.39)} \\ \hline \hline
\multicolumn{1}{|c|}{}                         & Joint         & \multicolumn{1}{c|}{63.78 (0.22)}                        & \multicolumn{2}{c|}{71.91 (0.57)}                                                \\ \cline{2-5} 
\multicolumn{1}{|c|}{}                         & FT            & \multicolumn{1}{c|}{39.48 (1.00)}                        & \multicolumn{1}{c|}{56.11 (0.57)}                        & 46.66 (0.59)          \\ \cline{2-5} 
\multicolumn{1}{|c|}{}                         & +CaSSLe       & \multicolumn{1}{c|}{53.02 (0.47)}                        & \multicolumn{1}{c|}{62.53 (0.11)}                        & 54.55 (0.12)          \\ \cline{2-5} 
\multicolumn{1}{|c|}{\multirow{-4}{*}{SimCLR}} & {+PNR} & \multicolumn{1}{c|}{{\textbf{53.47} (0.33)}}               & \multicolumn{1}{c|}{{\textbf{62.88} (0.19)}}               & {\textbf{54.79} (0.11)} \\ \hline \hline
\multicolumn{1}{|c|}{}                         & Joint         & \multicolumn{1}{c|}{69.36 (0.21)}                        & \multicolumn{2}{c|}{75.89 (0.22)}                                                \\ \cline{2-5} 
\multicolumn{1}{|c|}{}                         & FT            & \multicolumn{1}{c|}{49.46 (0.67)}                        & \multicolumn{1}{c|}{60.27 (0.27)}                        & 51.83 (0.46)          \\ \cline{2-5} 
\multicolumn{1}{|c|}{}                         & +CaSSLe       & \multicolumn{1}{c|}{54.46 (0.24)}                        & \multicolumn{1}{c|}{{64.98 (0.79)}} & 56.27 (0.63)          \\ \cline{2-5} 
\multicolumn{1}{|c|}{\multirow{-4}{*}{Barlow}} & {+PNR} & \multicolumn{1}{c|}{{\textbf{54.69} (0.21)}}               & \multicolumn{1}{c|}{{\textbf{65.38} (0.81)}}               & {\textbf{56.54} (0.42)} \\ \hline \hline
\multicolumn{1}{|c|}{}                         & Joint         & \multicolumn{1}{c|}{68.99 (0.28)}                        & \multicolumn{2}{c|}{75.52 (0.17)}                                                \\ \cline{2-5} 
\multicolumn{1}{|c|}{}                         & FT            & \multicolumn{1}{c|}{46.13 (0.88)}                        & \multicolumn{1}{c|}{60.77 (0.62)}                        & 51.04 (0.52)          \\ \cline{2-5} 
\multicolumn{1}{|c|}{}                         & +CaSSLe       & \multicolumn{1}{c|}{{57.36 (0.86)}} & \multicolumn{1}{c|}{62.31 (0.09)}                        & 57.47 (0.75)          \\ \cline{2-5} 
\multicolumn{1}{|c|}{\multirow{-4}{*}{BYOL}}   & {+PNR} & \multicolumn{1}{c|}{{\textbf{59.29} (0.25)}}               & \multicolumn{1}{c|}{{\textbf{64.23} (0.37)}}               & {\textbf{60.11} (0.91)} \\ \hline \hline
\multicolumn{1}{|c|}{}                         & Joint         & \multicolumn{1}{c|}{68.01 (0.36)}                        & \multicolumn{2}{c|}{75.08}                                                       \\ \cline{2-5} 
\multicolumn{1}{|c|}{}                         & FT            & \multicolumn{1}{c|}{46.88 (0.28)}                        & \multicolumn{1}{c|}{55.58 (0.22)}                        &       \multicolumn{1}{c|}{46.88 (0.34)}                 \\ \cline{2-5} 
\multicolumn{1}{|c|}{}                         & +CaSSLe       & \multicolumn{1}{c|}{47.76 (0.46)}                        & \multicolumn{1}{c|}{59.18 (0.36)}                        & 49.98 (0.53)          \\ \cline{2-5} 
\multicolumn{1}{|c|}{\multirow{-4}{*}{VICReg}} & {+PNR} & \multicolumn{1}{c|}{{\textbf{49.56} (0.56)}}               & \multicolumn{1}{c|}{{\textbf{62.48} (0.44)}}               & {\textbf{51.82} (0.50)} \\ \hline
\end{tabular}\label{table:class_il}
}
\vspace{-.1in}
\end{table}

\textbf{\noindent{Class-IL}} \ \
Table \ref{table:class_il} presents the experimental results of Class-IL with the CIFAR-100 and ImageNet-100 datasets. The results for Class-IL with CIFAR-100 (5T) can be found in Table \ref{fig:baseline_cifar100}.
In comparison to the state-of-the-art method, CaSSLe, PNR demonstrates superior performance across all scenarios. Notably, the combination of PNR with MoCo, BYOL, and VICReg exhibits significantly improved performance compared to their combination with CaSSLe. 
For example, in the ImageNet-100 experiments, "MoCo + PNR," "BYOL + PNR," and "VICReg + PNR" achieve a substantial gain of approximately 2-6\%, surpassing their combination with CaSSLe. 
As a result, we can confirm that the combination of PNR with each SSL method achieves state-of-the-art performance in the Class-IL scenarios.

% We believe that these findings underscore the effectiveness of the proposed PNR in Class-IL scenarios.
% Compared to the previous state-of-the-art method, CaSSLe, PNR achieves superior performance in all scenarios.
% Especially, the combination of PNR with MoCo, BYOL and VICReg achieves significantly improved performance compared to their combination with CaSSLe. For instance, in the ImageNet-100 experiments, "MoCo + PNR", "BYOL + PNR" and "VICReg + PNR" achieve a substantial gain of approximately 2-6\%, surpassing their combination with CaSSLe and establishing themselves as the new state-of-the-art methods.
% We believe that these results demonstrate the effectiveness of the proposed PNR in the Class-IL scenarios.

% \begin{wraptable}{r}{5.cm}
\begin{table}[h]
\vspace{-.1in}
\caption{{The experimental results of Class-IL with ImageNet-1k. The \textbf{bolded} result represents the best performance and we report results for a single seed.}}
% \vspace{-.15in}
\centering
\smallskip\noindent
% \vspace{-.1in}
\resizebox{.73\linewidth}{!}{
\begin{tabular}{c||cc|cc}
\hline
                          & \multicolumn{2}{c|}{MoCo}                                                   & \multicolumn{2}{c}{BYOL}                                                   \\ \cline{2-5} 
\multirow{-2}{*}{A$_{T}$} & \multicolumn{1}{c|}{5T}                                    & 10T            & \multicolumn{1}{c|}{5T}                                    & 10T            \\ \hline \hline
Joint                     & \multicolumn{2}{c|}{60.62}                                                  & \multicolumn{2}{c}{69.46}                                                  \\ \hline
FT                        & \multicolumn{1}{c|}{48.33}                                 & 42.55          & \multicolumn{1}{c|}{60.76}                                 & 57.15          \\ \hline
+CaSSLe                   & \multicolumn{1}{c|}{50.57}                                 & 43.38          & \multicolumn{1}{c|}{{64.78}}          & 61.93          \\ \hline
+PNR                      & \multicolumn{1}{c|}{\textbf{56.87}} & \textbf{55.88} & \multicolumn{1}{c|}{ \textbf{66.12}} & \textbf{62.56} \\ \hline
\end{tabular}\label{table:class_il_imagenet_1000}
}
\vspace{-.1in}
% \end{wraptable}
\end{table}

\textbf{\noindent{Class-IL with ImageNet-1k}} \ \
Table \ref{table:class_il_imagenet_1000} presents 
the experimental results of the ImageNet-1k dataset in Class-IL (5T and 10T).
We only conduct experiments for MoCo and BYOL, both of which have shown superior performance in previous experiments. 
We train a model for each method using the same hyperparameters employed in Class-IL with the ImageNet-100 dataset.
The experimental results in the table highlight the notable performance improvement of PNR in CSSL using the large-scale dataset, evident in both contrastive (MoCo) and non-contrastive (BYOL) learning methods. 
% These experimental results not only demonstrate the successful application of our proposed PNR to CSSL using both contrastive and non-contrastive learning methods again but also show its successful application even when dealing with a large-scale dataset.

\begin{table}[h]
% \vspace{-.15in}
\caption{{The experimental results of Data and Domain-IL. All results are reproduced performance. The values in parentheses indicate the standard deviation and the \textbf{bolded} result represents the best performance.}}
% \vspace{-.15in}
\centering
\smallskip\noindent
% \vspace{-.1in}
\resizebox{.98\linewidth}{!}{
\begin{tabular}{|cc||cc|c|}
\hline
\multicolumn{2}{|c||}{}                                         & \multicolumn{2}{c|}{Data-IL}                                                                                                          & Domain-IL                                                         \\ \cline{3-5} 
\multicolumn{2}{|c||}{}                                         & \multicolumn{2}{c|}{ImageNet-100}                                                                                                     & DomainNet                                                         \\ \cline{3-5} 
\multicolumn{2}{|c||}{\multirow{-3}{*}{A$_{T}$}}                & \multicolumn{1}{c|}{5T}                                           & 10T                                                               & 6T                                                                \\ \hline \hline
\multicolumn{1}{|c|}{}                         & Joint         & \multicolumn{2}{c|}{76.67 (0.56)}                                                                                                     & 48.20 (0.30)                                                      \\ \cline{2-5} 
\multicolumn{1}{|c|}{}                         & FT            & \multicolumn{1}{c|}{65.51 (0.72)}                                 & 60.50 (0.92)                                                      & 36.48 (1.01)                                                      \\ \cline{2-5} 
\multicolumn{1}{|c|}{}                         & +CaSSLe       & \multicolumn{1}{c|}{66.88 (0.32)}                                 & 59.72 (0.61)                                                      & 38.04 (0.24)                                                      \\ \cline{2-5} 
\multicolumn{1}{|c|}{\multirow{-4}{*}{MoCo}}   & {+PNR} & \multicolumn{1}{l|}{\textbf{69.98} (0.30)} & \multicolumn{1}{l|}{{{\textbf{67.83} (0.45)}}} & \multicolumn{1}{l|}{{\textbf{43.86} (0.17)}} \\ \hline
\multicolumn{1}{|c|}{}                         & Joint         & \multicolumn{2}{c|}{71.91 (0.57)}                                                                                                     & 48.50 (0.21)                                                      \\ \cline{2-5} 
\multicolumn{1}{|c|}{}                         & FT            & \multicolumn{1}{c|}{62.88 (0.30)}                                 & 56.47 (0.11)                                                      & 39.46 (0.20)                                                      \\ \cline{2-5} 
\multicolumn{1}{|c|}{}                         & +CaSSLe       & \multicolumn{1}{c|}{66.05 (0.95)}                                 & 61.68 (0.38)                                                      & 45.96 (0.19)                                                      \\ \cline{2-5} 
\multicolumn{1}{|c|}{\multirow{-4}{*}{SimCLR}} & {+PNR} & \multicolumn{1}{c|}{{\textbf{66.93} (0.12)}}                        & {\textbf{62.04} (0.28)}                                             & {\textbf{46.37} (0.13)}                                             \\ \hline \hline
\multicolumn{1}{|c|}{}                         & Joint         & \multicolumn{2}{c|}{75.89 (0.22)}                                                                                                     &            49.50 (0.32)                                                       \\ \cline{2-5} 
\multicolumn{1}{|c|}{}                         & FT            & \multicolumn{1}{c|}{66.47 (0.24)}                                 & 59.48 (1.33)                                                      & 41.87 (0.17)                                                      \\ \cline{2-5} 
\multicolumn{1}{|c|}{}                         & +CaSSLe       & \multicolumn{1}{c|}{{69.24 (0.36)}}          & {63.12 (0.28)}                               & 48.49 (0.04)                                                      \\ \cline{2-5} 
\multicolumn{1}{|c|}{\multirow{-4}{*}{Barlow}} & {+PNR} & \multicolumn{1}{c|}{{\textbf{70.16} (0.40)}}                        & {\textbf{64.06} (0.33)}                                                         & {\textbf{48.90} (0.07)}                                                         \\ \hline \hline
\multicolumn{1}{|c|}{}                         & Joint         & \multicolumn{2}{c|}{75.52 (0.17)}                                                                                                     &          53.80 (0.24)                                                         \\ \cline{2-5} 
\multicolumn{1}{|c|}{}                         & FT            & \multicolumn{1}{c|}{69.76 (0.45)}                                 & 61.39 (0.44)                                                      & 47.29 (0.08)                                                      \\ \cline{2-5} 
\multicolumn{1}{|c|}{}                         & +CaSSLe       & \multicolumn{1}{c|}{{\textbf{66.22} (0.13)}}          & {\textbf{63.33} (0.19)}                               & { 51.52 (0.18)}                               \\ \cline{2-5} 
\multicolumn{1}{|c|}{\multirow{-4}{*}{BYOL}}   & {+PNR} & \multicolumn{1}{l|}{{ {66.08 (0.15)}}} & \multicolumn{1}{l|}{{{63.10 (0.28)}}} & \multicolumn{1}{l|}{{ {\textbf{51.96} (0.08)}}} \\ \hline \hline
\multicolumn{1}{|c|}{}                         & Joint         & \multicolumn{2}{c|}{75.08 (0.14)}                                                                                                     &                 52.12 (0.17)                                                  \\ \cline{2-5} 
\multicolumn{1}{|c|}{}                         & FT            & \multicolumn{1}{c|}{64.02 (0.11)}                                 &      \multicolumn{1}{c|}{57.30 (0.09)}                                                             &   46.11 (0.14)                                                                \\ \cline{2-5} 
\multicolumn{1}{|c|}{}                         & +CaSSLe       & \multicolumn{1}{c|}{67.18 (0.15)}                                 & 61.50 (0.21)                                                      & 48.82 (0.12)                                                      \\ \cline{2-5} 
\multicolumn{1}{|c|}{\multirow{-4}{*}{VICReg}} & {+PNR} & \multicolumn{1}{c|}{{\textbf{67.68} (0.12)}}                        & {\textbf{62.08} (0.20)}                                             & {\textbf{48.95} (0.15)}                                             \\ \hline
\end{tabular}\label{table:data_il}
}
\vspace{-.05in}
\end{table}

\textbf{\noindent{Data-IL and Domain-IL}} \ \
Table \ref{table:data_il} presents the experimental results in Data- and Domain-IL using the ImageNet-100 dataset. In the results of Data-IL, the combination with PNR consistently achieves state-of-the-art performance in most cases. 
It is noteworthy that our PNR successfully integrates with MoCo once again, showcasing superior or nearly superior performance compared to other SSL methods at 5T and 10T. 
For instance, the combination "MoCo + PNR" demonstrates a 3-8\% performance enhancement compared to its pairing with CaSSLe. 
A similar trend is observed in Domain-IL, where the combination of CaSSLe with SSL methods already demonstrates superior performance close to their respective Joint's performance. However, PNR surpasses this by achieving additional performance improvements, thus setting new standards for state-of-the-art performance.
In contrast, "BYOL + PNR" performs worse than "BYOL + CaSSLe" in the Data-IL scenario. This can be attributed to the unique characteristics of Data-IL, which include shuffling and evenly distributing the ImageNet-100 dataset among tasks, leading to minimal distribution disparities between them. 
Additional discussion on this topic is available in Section \ref{sec:suboptimal} of the Appendix.

Additionally, we conducted all experiments using CaSSLe's code and were able to reproduce the CaSSLe's reported performance on CIFAR-100. However, despite various efforts, we could only achieve performance that was 4-6\% lower than the CaSSLe's reported performance in experiments using ImageNet-100 and DomainNet. More discussion regarding this matter is provided in Appendix \ref{sec:reproduce}.

\begin{figure*}[t]
\vspace{-.1in}
\centering 
\subfigure[MoCo + PNR]
{\includegraphics[width=0.32\linewidth]{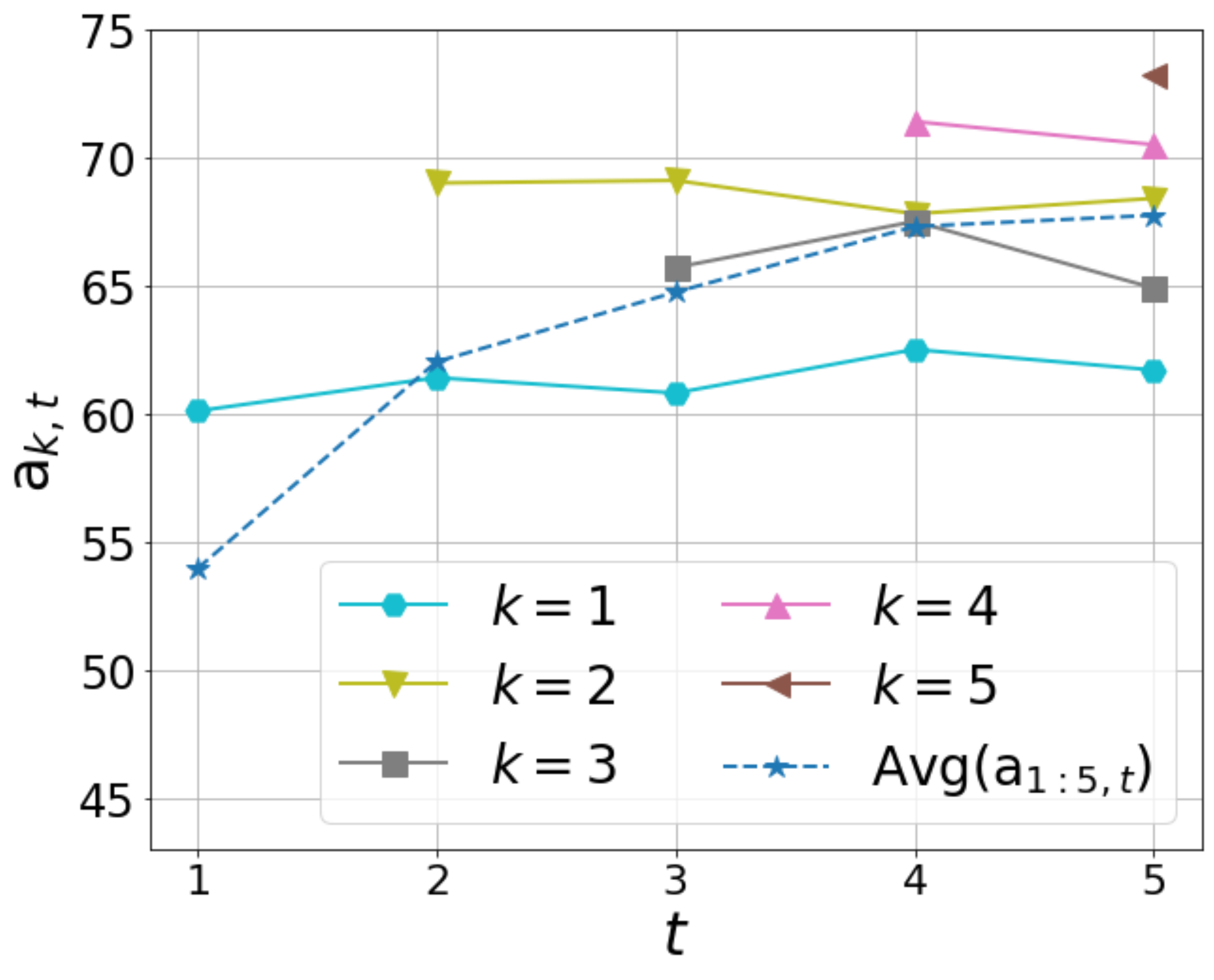}\label{figure:analysis_moco_pnr}}
\subfigure[{MoCo + CaSSLe}]
{\includegraphics[width=0.32\linewidth]{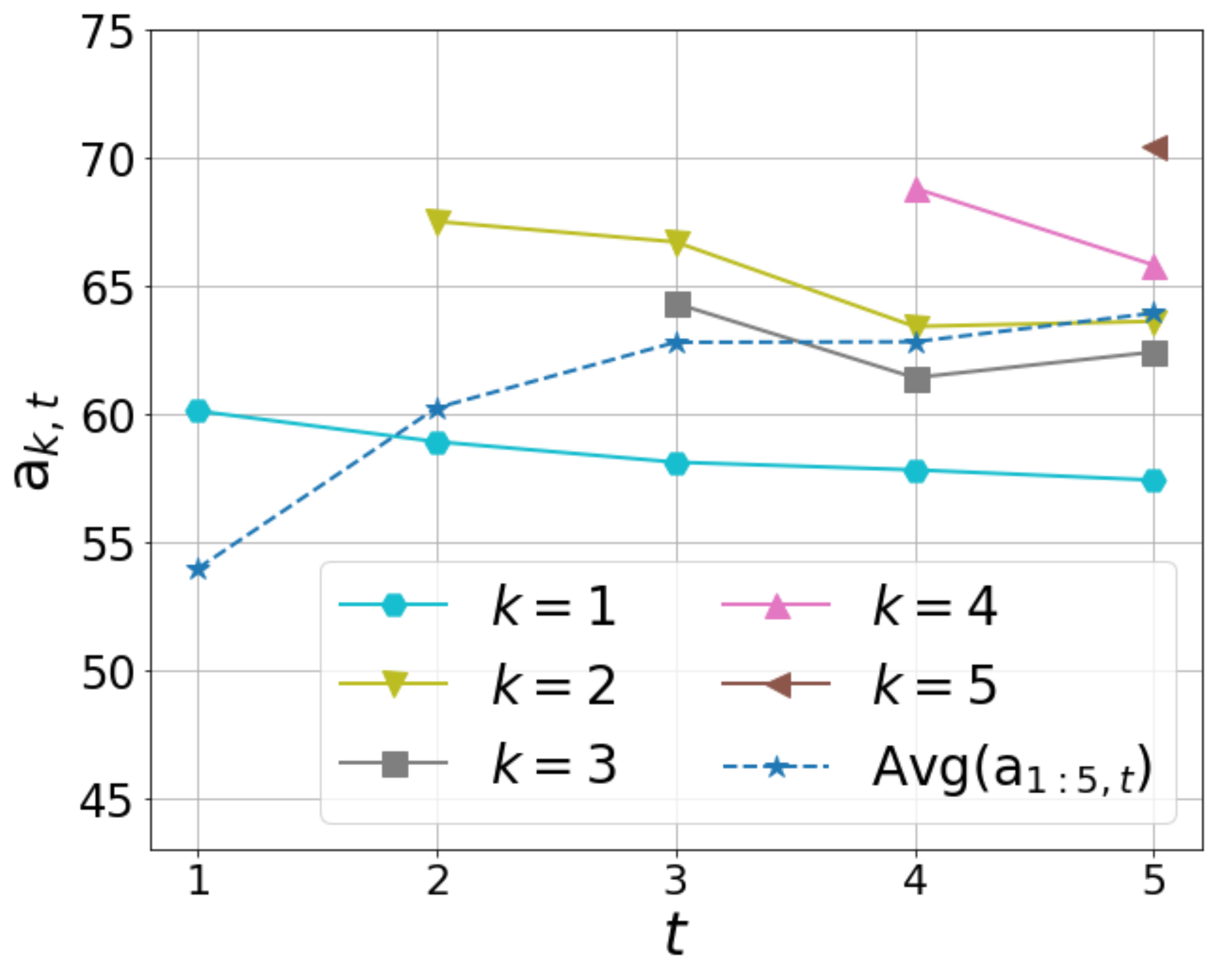}\label{figure:analysis_moco_cassle}}
\subfigure[{MoCo + FT}]
{\includegraphics[width=0.32\linewidth]{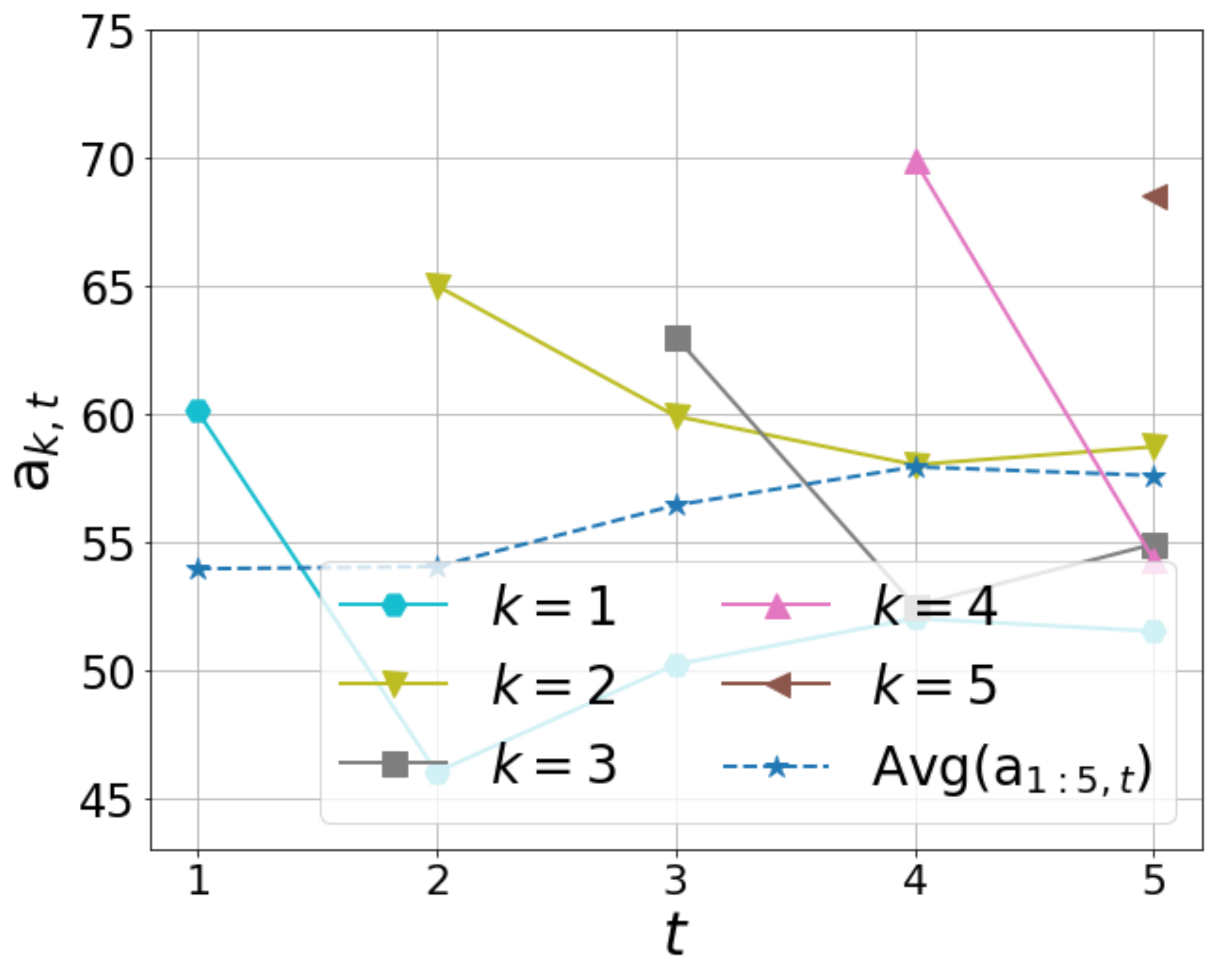}\label{figure:analysis_moco_ft}}
\subfigure[BYOL + PNR]
{\includegraphics[width=0.32\linewidth]{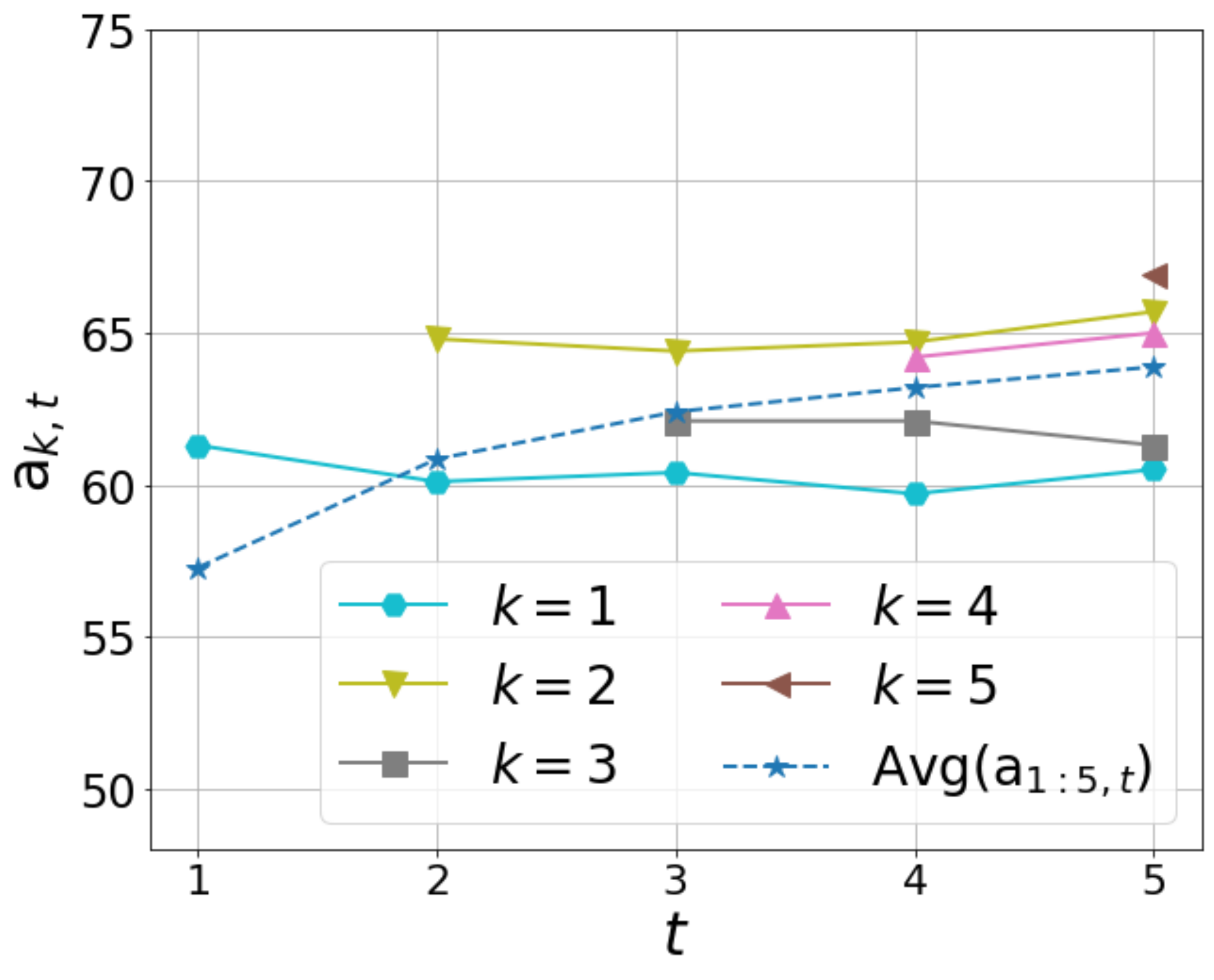}\label{figure:analysis_byol_pnr}}
\subfigure[{BYOL + CaSSLe}]
{\includegraphics[width=0.32\linewidth]{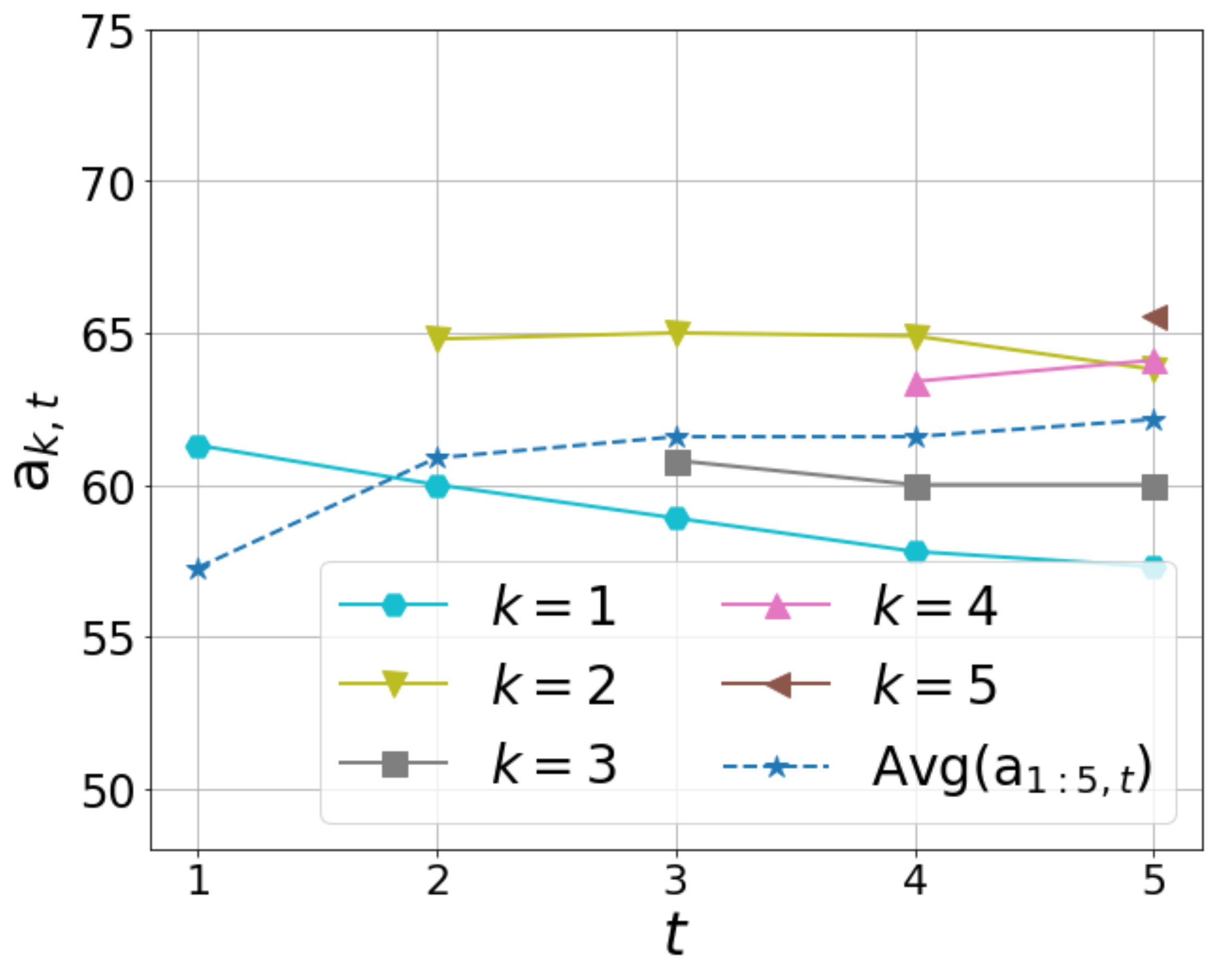}\label{figure:analysis_byol_cassle}}
\subfigure[{BYOL + FT}]
{\includegraphics[width=0.32\linewidth]{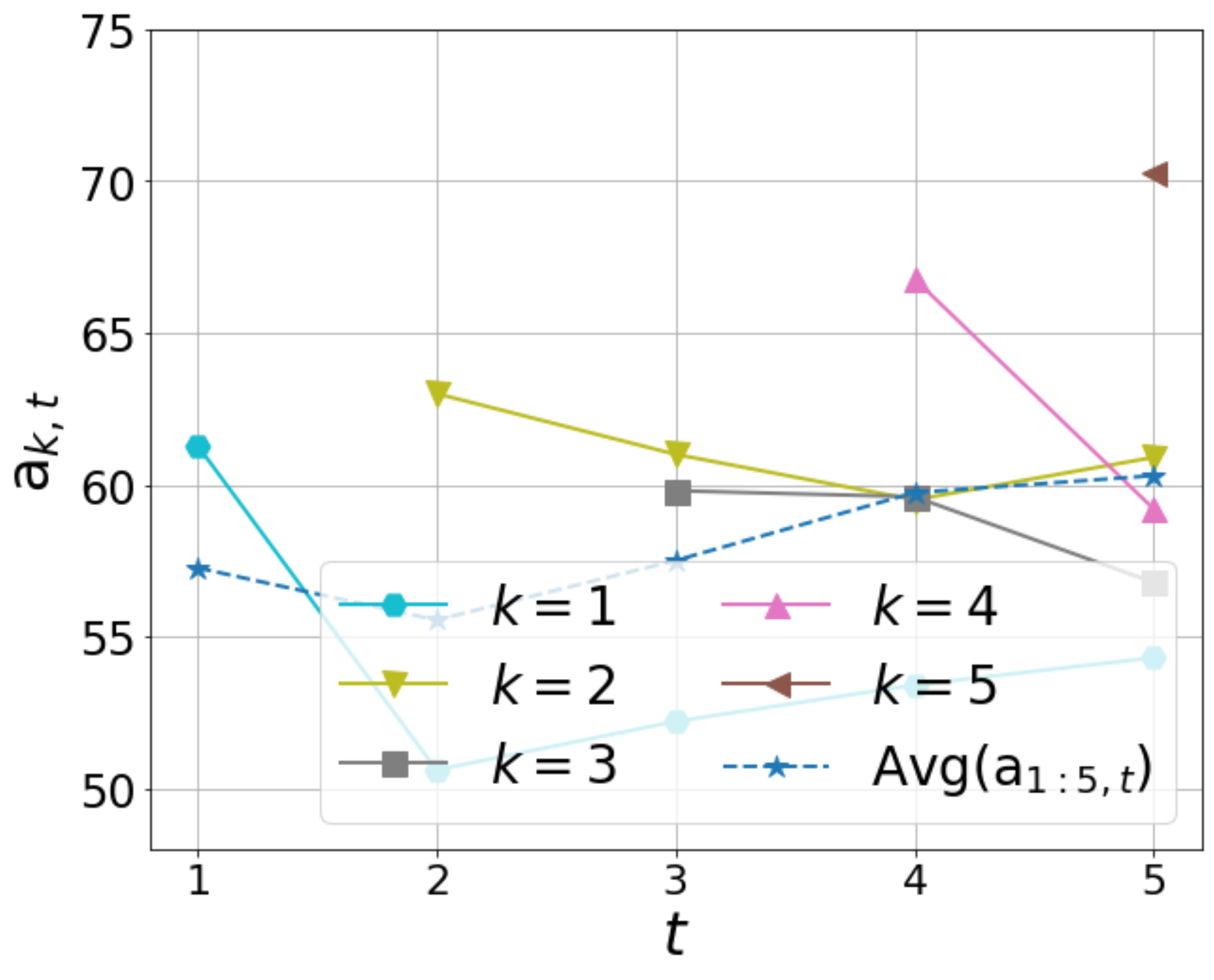}\label{figure:analysis_byol_ft}}
\vspace{-.1in}
\caption{{The graph illustrates the values of a$_{k,t}$ of each algorithm in the Class-IL (5T) scenario using the ImageNet-100 dataset. The measured stability ($S \downarrow$) and plasticity ($P \uparrow$) of each method are as follows: (a) $(S, P)=(1.23, 3.47)$, (b) $(S, P)=(2.80, 2.52)$, (c) $(S, P)=(3.13, 2.38)$, (d) $(S, P)=(0.4, -0.07)$, (e) $(S, P)=(1.5, -0.47)$, (f) $(S, P)=(4.9, 1.6)$.}}
\label{figure:each_task_graph}
\vspace{-.15in}
 \end{figure*}

\subsection{Experimental Analysis}

\textbf{\noindent{Analysis on plasticity and stability}} \ \ 
{Figure \ref{figure:each_task_graph} presents the experimental results of Class-IL (5T) using the ImageNet-100 dataset, showcasing graphs of {a$_{k,t}$} and {Avg(a$_{1:5,t}$)}. 
From Figure \ref{figure:analysis_moco_pnr} showing the result of "MoCo + PNR", we observe a general upward trend in a$_{k,t}$ across all tasks. Remarkably, the performance of the initial task (a$_{k=1,t}$) remains relatively stable and even exhibits slight improvement as subsequent tasks are learned. 
Conversely, the results depicted in Figures \ref{figure:analysis_moco_cassle} and \ref{figure:analysis_moco_ft} of "MoCo + CaSSLe" and "MoCo + FT" indicate that, while their Avg(a$_{1:5,t}$) gradually increases, certain task performances experience gradual declines (e.g., a$_{k=2,t}$ of "MoCo + CaSSLe" and most $k$ of "MoCo + FT"), showing suffering from catastrophic forgetting than "MoCo + PNR".
Furthermore, the numerical assessments of plasticity ($P$) and stability ($S$) of each algorithm, as presented in the caption of Figure \ref{figure:each_task_graph}, demonstrate that "MoCo + PNR" achieves its performance improvement through superior \textit{plasticity} and \textit{stability} compared to other baselines. 

% Figure \ref{figure:analysis_byol_pnr}, \ref{figure:analysis_byol_cassle} and \ref{figure:analysis_byol_ft} depict the results of the experimental analysis conducted on "PNR + BYOL", "BYOL + CaSSLe", and "BYOL + FT" in Class-IL (5T) experiments using the ImageNet-100 dataset. The trends observed here align with previous results. First, there is a gradual increase in their Avg(a$_{1:5,t}$). However, it is noteworthy that a$_{k=1,t}$, a$_{k=3,t}$ for "BYOL + CaSSLe" decrease across tasks. Similarly, "BYOL + FT" shows a degraded performance in most $k$. In contrast, the application of the proposed PNR not only maintains a$_{k=1,t}$ well but also results in a gradual increase in a$_{k=2,t}$ and a$_{k=4,t}$, suggesting that PNR surpasses "BYOL + CaSSLe" in terms of plasticity and stability. Also, The plasticity ($P$) and stability ($S$) measurements mentioned in the caption of Figure \ref{figure:each_task_graph} further substantiate these experimental results.
% These analyses not only underscore the effectiveness of incorporating the pseudo-negative but also provide additional evidence of our PNR's superior performance. 
% The corresponding analysis for other baselines is available in Section \ref{sec:analysis} of the Appendix.

Figures \ref{figure:analysis_byol_pnr}, \ref{figure:analysis_byol_cassle}, and \ref{figure:analysis_byol_ft} illustrate the results of our experimental analysis conducted on "PNR + BYOL," "BYOL + CaSSLe," and "BYOL + FT" in Class-IL (5T) experiments using the ImageNet-100 dataset. These findings align with previous results, showing a gradual increase in Avg(a$_{1:5,t}$). 
However, significant declines are seen in a$_{k=1,t}$ and a$_{k=3,t}$ for "BYOL + CaSSLe" across tasks, whereas "BYOL + FT" shows decreased performance in most $k$. In contrast, the application of our proposed PNR not only maintains a$_{k=1,t}$ effectively but also leads to a gradual increase in a$_{k=2,t}$ and a$_{k=4,t}$, indicating that our PNR outperforms "BYOL + CaSSLe" in terms of plasticity and stability. Additionally, the plasticity ($P$) and stability ($S$) measurements mentioned in the caption of Figure \ref{figure:each_task_graph} further support these experimental findings.

In conclusion, these analyses not only highlight the effectiveness of integrating the pseudo-negative but also provide further evidence of the superior performance of our PNR. Additional analysis for other baselines can be found in Section \ref{sec:analysis} of the Appendix.

% Figure \ref{figure:each_task_graph2} depicts the results of the experimental analysis conducted on "PNR + BYOL", "BYOL + CaSSLe", and "BYOL + FT" in Class-IL (5T) experiments using the ImageNet-100 dataset. The trends observed here align with those presented in the manuscript. Initially, there is a gradual increase in their Avg(a$_{1:5,t}$). However, it is noteworthy that a$_{k=1,t}$, a$_{k=3,t}$ for "BYOL + CaSSLe" decrease across tasks. Similarly, "BYOL + FT" experiences a decline in most $k$. In contrast, the application of the proposed PNR not only maintains a$_{k=1,t}$ but also results in a gradual increase in a$_{k=2,t}$ and a$_{k=4,t}$, suggesting that PNR surpasses "BYOL + CaSSLe" in terms of plasticity and stability. 

\textbf{\noindent{Analysis on the impact of pseudo-negatives}} \ \ Table \ref{table:queue_main} presents the experimental results for different queue sizes (\textit{i.e.}, the size of $\mathcal{PN}_1(i)$ and $\mathcal{PN}_2(i)$) in "MoCo + PNR". Note that the default queue size is 65536 for all previous experiments. We observe that performance remains relatively consistent when the queue size exceeds 16384. However, significant performance degradation is evident with reduced queue sizes (\textit{i.e.}, reduced pseudo-negatives), particularly at 256. Based on this result, we affirm that the superior performance of "MoCo + PNR" is attributed to the utilization of a large number of pseudo-negatives.

\begin{table}[h]
\vspace{-.1in}
\caption{Experiments with different queue sizes.}
\label{table:queue_main}
\centering
\smallskip\noindent
% \vspace{-.2in}
\resizebox{.98\linewidth}{!}{
\begin{tabular}{|c||c|c|c|c|c|c|}
\hline
Queue size & 256   & 512  & 2048  & 16384 & {{65536}} & 131072 \\ \hline \hline
A$_{5}$    & 61.03 & 61.68 & 61.90 & 62.10 & {{62.36}} & 62.01  \\ \hline
\end{tabular}
}
\vspace{-.05in}
\end{table}

\begin{table*}[t]
\vspace{-.1in}
\caption{{Experimental results of semi-supervised learning and downstream tasks. "10\%" and "1\%" denote the percent of used supervised datasets for semi-supervised learning. Also, "CIL", "DIL", "DTs" and 
"Cli." means "Class-IL", "Data-IL", "Downstream Tasks" and "Clipart", respectively. The \textbf{bolded} result represents the best performance}}
% \vspace{-.15in}
\label{table:semi_supervised_learning}
\centering
\smallskip\noindent
% \vspace{-.1in}
\resizebox{.98\linewidth}{!}{
\begin{tabular}{|cc||c|c|c|c||c|c|}
\hline
\multicolumn{2}{|c|}{}                                & \begin{tabular}[c]{@{}c@{}}MoCo  + CaSSLe\end{tabular} & \begin{tabular}[c]{@{}c@{}}SimCLR  + CaSSLe\end{tabular} & \begin{tabular}[c]{@{}c@{}}Barlow  + CaSSLe\end{tabular} & \begin{tabular}[c]{@{}c@{}}BYOL  + CaSSLe\end{tabular} & {\begin{tabular}[c]{@{}c@{}}MoCo + PNR\end{tabular}} & {\begin{tabular}[c]{@{}c@{}}BYOL + PNR\end{tabular}} \\ \hline \hline
\multicolumn{1}{|c|}{}                          & CIL & {56.48}                             & {55.16}                               & {55.10}                           & {54.22}                             & {{\textbf{61.74}}}                         & {{57.36}}                         \\ \cline{2-8} 
\multicolumn{1}{|c|}{\multirow{-2}{*}{\rotatebox{90}{10\%}}}    & DIL & {62.66}                             & {60.14}                               & {62.18}                           & {59.48}                             & {{\textbf{65.28}}}                         & {{58.04}}                         \\ \hline 
\multicolumn{1}{|c|}{}                          & CIL & {39.14}                             & {40.86}                               & {41.90}                           & {36.86}                             & {{\textbf{46.48}}}                         & {{40.82}}                         \\ \cline{2-8} 
\multicolumn{1}{|c|}{\multirow{-2}{*}{\rotatebox{90}{1\%}}}     & DIL & {51.04}                             & {49.72}                               & {54.10}                           & {46.06}                             & {{\textbf{54.88}}}                         & {{44.46}}                         \\ \hline \hline
\multicolumn{1}{|c|}{}                          & CIL & {58.61}                             & {56.73}                               & {58.13}                           & {61.35}                             & {{62.04}}                         & {{\textbf{62.53}}}                         \\ \cline{2-8} 
\multicolumn{1}{|c|}{\multirow{-2}{*}{\rotatebox{90}{DTs}}}   & DIL & {59.15}                             & {56.97}                               & {59.54}                           & {61.61}                             & {{\textbf{62.35}}}                         & {{61.95}}                         \\ \hline
\multicolumn{1}{|c|}{}                          & CIL & 28.32                                                    & 34.68                                                      & 37.42                                                  & 38.98                                                    & {38.86}                                                & {\textbf{41.57}}                                                \\ \cline{2-8} 
\multicolumn{1}{|c|}{\multirow{-2}{*}{\rotatebox{90}{Cli.}}} & DIL & 29.74                                                    & 34.17                                                      & 36.13                                                  & 37.04                                                    & {38.33}                                                & {\textbf{40.06}}                                                \\ \hline
\end{tabular}
}
\vspace{-.1in}
\end{table*}

\textbf{\noindent{Semi-supervised learning and downstream tasks}} \ \
To evaluate the quality of learned representations in a more diverse way, we conduct experiments in a semi-supervised scenario. Specifically, we consider a scenario where a linear classifier is only trained using only 1\% or 10\% of the entire supervised ImageNet-100 dataset. We evaluate each encoder trained in the Class-IL (5T) and Data-IL (5T) using the ImageNet-100 dataset. The experimental results are presented in the upper rows of Table \ref{table:semi_supervised_learning}. 
Notably, when compared to CaSSLe, applying PNR to both MoCo and BYOL yields approximately 3-6\% performance improvements in both 10\% and 1\%, showing new state-of-the-art performances.
The lower rows of Table \ref{table:semi_supervised_learning} present the results of linear evaluation for downstream tasks conducted on the same encoders. For the three datasets, we report the average accuracy of linear evaluation results on STL-10~\citep{(stl10)coates2011analysis}, CIFAR-10, and CIFAR-100~\citep{(cifar)krizhevsky2009learning} datasets. Additionally, we set Clipart within the DomainNet~\citep{(domainnnet)peng2019moment} dataset as the downstream task and report the results of linear evaluation using it. From these experimental results, we once again demonstrate that both "MoCo + PNR" and "BYOL + PNR" achieve performance improvements compared to the CaSSLe's performance. 

More detailed results and additional findings (e.g., experiments using the model trained in Class-IL (10T) and Data-IL (10T) are provided in Section \ref{sec:downstream} of the Appendix.

\textbf{\noindent{Computational cost}} \ \ 
Note both PNR and CaSSLe incur almost identical computational costs, except for "MoCo + PNR," which involves an additional queue to store negatives from the $t-1$ model. However, the memory size required for this additional queue is negligible.

\textbf{\noindent{Ablation study}} \ \
Table \ref{table:ablation_study} presents the results of the ablation study conducted on CIFAR-100 in the Class-IL (5T) scenario. The first row represents the performance of "MoCo + PNR" when all negatives are used. Cases 1 to 3 illustrate the results when one of pseudo-negatives, $\mathcal{PN}_1(i)$ and  $\mathcal{PN}_2(i)$ is excluded or when both are omitted.
The experimental results show that the absence of these negatives leads to a gradual decrease in performance, indicating that "MoCo + PNR" acquires superior representations by leveraging both the pseudo-negatives.
Specifically, when $\mathcal{PN}_2(i)$ is omitted, the performance degradation is more significant compared to when the negative from $\mathcal{PN}_1(i)$ is absent.
Moreover, the result of Case 4, where a queue size twice as large is used for "MoCo + CaSSLe", demonstrates that using the proposed pseudo-negatives is different from simply increasing the queue size. Note that ablation study for PNR with non-contrastive learning, such as BYOL and VICReg, can be confirmed by comparing the results of "+CaSSLe" and "+PNR" in Table \ref{fig:baseline_cifar100}, \ref{table:class_il}, \ref{table:class_il_imagenet_1000}, and \ref{table:data_il}.

\begin{table}[h]
\vspace{-.1in}
\caption{{Ablation study of "MoCo + PNR" in Class-IL (5T) using the ImageNet-100 dataset.}}
% \vspace{-.1in}
\centering
\smallskip\noindent
\resizebox{.98\linewidth}{!}{
\begin{tabular}{c||c|c|c|c||c}
\hline
       & \begin{tabular}[c]{@{}c@{}}$\mathcal{PN}_1(i)$\end{tabular} & \begin{tabular}[c]{@{}c@{}}$\mathcal{PN}_2(i)$\end{tabular} & + CaSSLe & $|$queue$|$ $\times 2$ & A$_{5}$ \\ \hline \hline
PNR &          \cmark                                 &             \cmark                                  &      \xmark    &          \xmark          & 62.36        \\ \hline
Case 1 &           \xmark                                &                    \cmark                           &      \xmark    &     \xmark               & 61.26   \\ \hline
Case 2 &              \cmark                             &             \xmark                                  &     \xmark     &      \xmark              & 60.92   \\ \hline
Case 3 &                                 \xmark          &                             \xmark                  &       \xmark   &            \xmark        & 59.97   \\ \hline
Case 4 &                         \xmark                  &           \xmark                                    &    \cmark      &            \cmark        &  60.09  \\ \hline
\end{tabular}\label{table:ablation_study}
}
\vspace{-.in}
\end{table}

Moreover, we present further experimental results of applying PNR to supervised contrastive learning in Section \ref{sec:supcon} of the Appendix.

% \vspace{-.15in}
% \noindent{$\mathcal{L}_\mathrm{plasticity}$ with CasSSLe} \ \

% \noindent{Training cost} \ \

% \noindent{Representation analysis using CKA} \ \

\section{Limitation and Future Work}
There are several limitations to our work. First, we focused on Continual Self-Supervised Learning (CSSL) with CNN-based architectures (e.g., ResNet). However, we believe that our proposed idea and loss function could be applied to CSSL using vision transformer-based models. Second, we only considered CSSL in the computer vision domain. Nonetheless, we believe that the concept of pseudo-negatives could be extended to CSSL in other domains, such as natural language processing. We defer these explorations to the future work.

\section{Concluding Remarks}

We present Pseudo-Negative Regularization (PNR), a simple yet novel method employing pseudo-negatives in Continual Self-Supervised Learning (CSSL). First, we highlight the limitations of the traditional CSSL loss formulation, which may impede the learning of superior representations when training a new task. To overcome this challenge, we propose considering the pseudo-negatives generated from both previous and current models in CSSL using contrastive learning methods. Furthermore, we expand the concept of PNR to non-contrastive learning methods by incorporating additional regularization. Through extensive experiments, we confirm that our PNR not only can be applied to self-supervised learning methods, but also achieves state-of-the-art performance with superior stability and plasticity.

% {We introduce the Pseudo-Negative Regularization (PNR), a simple but novel approach using pseudo negative in continual self-supervised learning (CSSL). Initially, we highlight the limitations of the traditional CSSL loss formulation, which may hinder learning better representations during training a new task. To address this challenge, we propose considering pseudo negative from both previous and current models for InfoNCE-based contrastive learning. Furthermore, we show that the idea of PNR can be extended to non-contrastive methods by adding extra regularization. Through extensive experiments, we observe that not only our PNR can be applied to various self-supervised learning methods but also achieves state-of-the-art performance with superior stability and plasticity.}

\newpage

\section*{Impact Statement}

This paper presents a Pseudo-Negative Regularization (PNR) framework designed to advance Continual Self-Supervised Learning (CSSL). By balancing the trade-off between plasticity and stability during CSSL, our work contributes to the development of more energy-efficient AI systems, supporting both environmental sustainability and technological progress. Furthermore, our framework paves the way for more adaptive AI applications across various sectors.

\section*{Acknowledgment}
This work was supported in part by the National Research Foundation of Korea (NRF) grant [No.2021R1A2C2007884] and by Institute of Information \& communications Technology Planning \& Evaluation (IITP) grants
[RS-2021-II211343, RS-2021-II212068, RS-2022-II220113,
RS-2022-II220959] funded by the Korean government (MSIT). It was also supported by AOARD Grant No. FA2386-23-1-4079 and SNU-Naver Hyperscale AI Center. 
% While there are various potential societal consequences of our work, none require specific highlighting here.

% \csmcm{The work was supported in part by NRF grant [2021R1A2C2007884, 2021M3E5D2A01024795], IITP grant [No.2021-0-01343, No.2021-0-02068, No.2022-0-00113, No.2022-0-00959] funded by the Korean government, and SNU-Naver Hyperscale AI Center.}

% \newpage

\bibliography{reference}
\bibliographystyle{iclr2024_conference}
\newpage

\onecolumn
\appendix
% \section{Appendix}
\pdfoutput=1

\section{Supplementary Materials for Section 3}
\subsection{The Gradient Analysis}\label{sec:sup_gradient}

Here, we give the gradient analysis of our PNR loss for contrastive learning methods. For simplicity, we assume that $g(\cdot)$ is an identity map instead of an MLP layer, hence $g({\bm z}_{t,i}^A) = \bm z_{t,i}^A$. Now, we can show the gradient of $\mathcal{L}_{t}^{\text{CSSL}}$ with respect to $\bm z_{t,i}^A$ becomes
\begin{equation}
\frac{\partial (\frac{1}{2}\mathcal{L}_{t}^{\text{CSSL}})}{\partial \bm z_{t,i}^A} = -\underbrace{\Bigg(\frac{\bm z_{t,i}^B + \bm z_{t-1,i}^A}{2}\Bigg)}_{(a)} + \underbrace{\Bigg\{\sum_{\bm z_{j}\in \mathcal{N}_1(i)\cup \mathcal{PN}_1(i)} \bm z_{j} \cdot S^1_{t,i}(\bm z_{j}) + \sum_{\bm z_{j}\in \mathcal{N}_2(i)\cup \mathcal{PN}_2(i)} \bm z_{j} \cdot S^2_{t,i}(\bm z_{j})\Bigg\}}_{(b)}, \label{eqn:gradient}\ \ \ \nonumber
\end{equation}
in which $S^1_{t,i}(u) = \mathrm{exp}(\bm z_{t,i}^A \ \cdot \ u) / {\sum_{\bm z_{j}\in \mathcal{N}_1(i)\cup \mathcal{PN}_1(i)}{\mathrm{exp}(\bm z_{t,i}^A \cdot \bm z_{j} / \tau)}}$, $S^2_{t,i}(u) = \mathrm{exp}(\bm z_{t,i}^A \ \cdot \ u) / {\sum_{\bm z_{j}\in \mathcal{N}_2(i)\cup \mathcal{PN}_2(i)}{\mathrm{exp}(\bm z_{t,i}^A \cdot \bm z_{j} / \tau)}}$, and $\sum_{\bm z_{j}\in \mathcal{N}_1(i)\cup \mathcal{PN}_1(i)}{S^1_{i,t}(\bm z_{j})} + \sum_{\bm z_{j}\in \mathcal{N}_2(i)\cup \mathcal{PN}_2(i)}{S^2_{i,t}(\bm z_{j})} = 1$. 
Similarly as in the Unified Gradient of the InfoNCE loss \citep{(selfsup_analysis)tao2022exploring}, we can make the following interpretations. Namely, the negative gradient step can be decomposed into two parts, part (a) and the negative of part (b) above.

Part (a) is the average of the embedding of $h_{\bm \theta_t}$ for the positive sample and the embedding of $h_{\bm \theta_{t-1}}$ for the input sample (\textit{i.e., }$x_i$). Hence, this direction encourages the model to learn new representations while taking the \textit{stability} from the previous model into account. On the other hand, the negative of part (b) is the repelling direction from the center of mass point among the negative sample embeddings in $\mathcal{N}_1(i)\cup \mathcal{PN}_1(i)$ and $\mathcal{N}_2(i)\cup \mathcal{PN}_2(i)$, in which each element $u\in\mathcal{N}_1(i)\cup \mathcal{PN}_1(i)$ and $u\in\mathcal{N}_2(i)\cup \mathcal{PN}_2(i)$ has the probability mass $S^1_{t,i}(u)$ and $S^2_{t,i}(u)$, respectively. Thus, this direction promotes the new representations to be more discriminative from the current and previous models' negative sample embeddings, leading to improved \textit{plasticity}. 
Our gradient analysis allows us to better understand the graphical representation in the left figure of Figure \ref{figure:illustration}.

\subsection{{PNR Implementation for Non-Contrastive Learning Methods}}\label{sec:implementation}

\subsubsection{BYOL + PNR}

The implementation of "BYOL + PNR" following Equation (\ref{eqn:noncont_l2_1}) and (\ref{eqn:noncont_l2_2}) is as below:

\begin{align}
\mathcal{L}^{\text{SSL}^{*}}_1(\{\bm z_t^A, \bm z_t^B)\} = 
\|g_{\theta_t}(\bm z_t^A) - z_{\xi}^{\text{B}}\|^2_2\label{eqn:byol}
\end{align}

\begin{align}
\mathcal{L}^{\text{SSL}^{*}}_2(\{g(\bm z_t^A), \bm z_{t-1}^A, \bm z_{t-1}^B\}) &=  \|g(\bm z_t^A) - \bm z_{t-1}^A\|^2_2\label{eqn:byol_cassle} \\&- \lambda_{\text{PNR}}  
\|g(\bm z_t^A) - \bm z_{t-1}^B\|^2_2\label{eqn:byol_pnr}, 
\end{align}

where $q_{\theta_t}$ represents an MLP layer in BYOL, and $z_{\xi}^{\text{B}}$ corresponds to output features of the target network obtained through momentum updates using $\bm h_\theta$. $\|\cdot\|^2_2$ represents the squared $L_2$ norm and $\lambda_{\text{PNR}}$ is a hyperparameter.

\subsubsection{VICReg + PNR}

The implementation of "VICReg + PNR" following Equation (\ref{eqn:noncont_l2_1}) and (\ref{eqn:noncont_l2_2}) is as below:

\begin{align}
\mathcal{L}^{\text{SSL}^{*}}_1(\{\bm z_t^A, \bm z_t^B)\} = \lambda s(\bm z_t^A, \bm z_t^B) + \mu [v(\bm z_t^A) + v(\bm z_t^B)] + \nu [c(\bm z_t^A) + c(\bm z_t^B)] \label{eqn:vicreg}\\
\end{align}
\begin{align}
\mathcal{L}^{\text{SSL}^{*}}_2(\{g(\bm z_t^A), \bm z_{t-1}^A, \bm z_{t-1}^B\}) = 
\lambda_{\text{CaSSLe}} [s(g(\bm z_t^A), \bm z_{t-1}^A)*0.5]\label{eqn:cassle_vicreg} \\
- \lambda_{\text{PNR}} [s(g(\bm z_t^A), \bm z_{t-1}^B)*0.5]\label{eqn:PNR_vicreg},
\end{align}

 where Equation (\ref{eqn:vicreg}) is the original VICReg loss function and $s(\cdot,\cdot)$ denotes the mean-squared euclidean distance between each pair of vectors. $\lambda_{\text{CaSSLe}}$ and $\lambda_{\text{PNR}}$ are hyperparameters. 
 
 % Equation (\ref{eqn:cassle_vicreg}) represents CaSSLe's distillation, while Equation (\ref{eqn:PNR_vicreg}) serves as an additional regularization for PNR in non-contrastive learning, sharing a similar form with Equation (\ref{eqn:PNR_byol}) used in PNR(BYOL).

% \begin{align}
% \mathcal{L}^{t}_{\text{PNR, BYOL}}(x^{\text{A}}, x^{\text{B}}, \theta_{t}, \theta_{t-1}) = 
% \|g_{\theta_t}(h_{\theta_t}(x^{\text{A}})) - {z_{\xi}}(x^{\text{B}})\|^2_2\label{eqn:byol} \\ +  
% \|\text{Pred}(h_{\theta_t}(x^{\text{A}})) - h_{\theta_{t-1}}(x^{\text{A}})\|^2_2\label{eqn:cassle_byol} \\ -  
% \|\text{Pred}(h_{\theta_t}(x^{\text{A}})) - h_{\theta_{t-1}}(x^{\text{B}})\|^2_2\label{eqn:PNR_byol}, 
% \end{align}

\section{Supplementary Materials for Section 4}

\subsection{Measures for Stability and Plasticity}\label{sec:measure}

To evaluate each CSSL algorithm in terms of stability and plasticity, we use measures for them, following~\cite{(cassle)fini2022self, (cpr)cha2021cpr}, as shown in below:

\begin{compactitem}
    \item Stability: $S=\dfrac{1}{T-1}\sum_{i=1}^{T-1}\max_{t\in \{1, \dots, T\}} (a_{i,t} - a_{i,T})$
    \item Plasticity: $P=\dfrac{1}{T-1}\sum_{j=1}^{T-1}\dfrac{1}{T-j}\sum_{i=j+1}^{T}(a_{i,j} - FT_{i})$
\end{compactitem}

Here, $FT_{i}$ signifies the linear evaluation accuracy (on the validation dataset of task $i$) of the model trained with a SSL algorithm for task $i$.

\subsection{Discussion on Suboptimal Performance of CaSSLe and PNR in Data-IL}\label{sec:suboptimal}

As shown in Table \ref{table:data_il}, combining BYOL with CaSSLe and PNR in Data-IL led to suboptimal performance, particularly in the 5T scenario. This can be attributed to the unique characteristics of Data-IL, briefly discussed in the CaSSLe paper. Data-IL involves shuffling and evenly distributing the ImageNet-100 dataset among tasks, resulting in minimal distribution disparities between them. During BYOL training (using Equation (\ref{eqn:byol})), the target encoder ($\xi$) retains some information from the current task which is conceptually similar to the previous task, due to momentum updates from the training encoder ($\bm h_{\theta_t}$). {As a result, fine-tuning solely with BYOL can produce a robust Data-IL outcome due to the substantial similarity in distribution between tasks. However, the incorporation of CaSSLe (Equation (\ref{eqn:byol_cassle})) and PNR (Equation (\ref{eqn:byol_pnr})) may conflict with Equation (\ref{eqn:byol}) in Data-IL.}

{On the contrary, when applied to the Domain-IL scenario using the DomainNet dataset where distinct variations in input distribution are evident for each task, "BYOL + PNR" effectively demonstrates the feasibility of augmenting negative representations  as outlined in Table \ref{table:data_il}. This reinforces our conviction that the suboptimal results observed in Data-IL are solely attributable to the unique and artificial circumstances inherent to Data-IL.}

% We use $S=\dfrac{1}{T-1}\sum_{i=1}^{T-1}\max_{t\in \{1, \dots, T\}} (a_{i,t} - a_{i,T})$ and $P=\dfrac{1}{T-1}\sum_{j=1}^{T-1}\dfrac{1}{T-j}\sum_{i=j+1}^{T}(a_{i,j} - FT_{i})$, as the measure of stability and plasticity, respectively.

\subsection{{Experiments to Reproduce the CaSSLe's Reported Performance}}\label{sec:reproduce}

We conduct all experiments based on the code provided by CaSSLe. For experiments on CIFAR-100, we are able to achieve results similar to those reported in the CaSSLe paper. However, despite our efforts to closely replicate the environment, including utilizing the packages and settings provided by CaSSLe, we consistently obtain results approximately 2-6\% lower in experiments on ImageNet-100 and DomainNet. This issue has been raised and discussed on the issue page of CaSSLe's GitHub repository (\href{https://github.com/DonkeyShot21/cassle/issues/12}{Link 1}, \href{https://github.com/DonkeyShot21/cassle/issues/10}{Link 2}), where researchers have consistently reported performance discrepancies ranging from 4-6\% lower than those stated in the CaSSLe paper, particularly in ImageNet-100 experiments.

Considering this discrepancy, we hypothesized that it might be related to the hyperparameter issue. Therefore, we conduct experiments by varying key hyperparameters used in linear evaluation and CSSL, such as Learning Rate (LR) and Mini-Batch Size (MBS), to address this concern. The results of these experiments are presented in Table \ref{table:hyperparameter1} and Table \ref{table:hyperparameter2}. In Table \ref{table:hyperparameter1}, we use the CSSL model trained with default hyperparameters for CSSL, while in Table \ref{table:hyperparameter2}, we conduct experiments using the default hyperparameters for linear evaluation. Despite exploring various hyperparameter settings, we are unable to achieve the performance reported in the CaSSLe paper. However, across all considered hyperparameter configurations, our proposed "MoCo + PNR" consistently outperforms "MoCo + CaSSLe," confirming its superior performance.

% \vspace{-.1in}
\begin{table}[t]
\caption{{Experimental results from various hyperparameters utilized in CSSL. Text highlighted in \textbf{bold} indicates the results obtained using default hyperparameters.}}
% \vspace{-.15in}
\label{table:hyperparameter1}
\centering
\smallskip\noindent
% \vspace{-.1in}
\resizebox{.5\linewidth}{!}{
\begin{tabular}{|c||c|c|c|}
\hline
\begin{tabular}[c]{@{}c@{}}Hyperparameter set \\ (for CSSL)\end{tabular} & FT             & MoCo + CaSSLe  & MoCo + PNR \\ \hline \hline
LR = 0.2, MBS = 128                                                      & 57.34          & 63.56          & 66.50      \\ \hline
\textbf{LR = 0.4, MBS = 128}                                             & \textbf{57.52} & \textbf{62.96} & 67.08      \\ \hline
LR = 0.8, MBS = 128                                                      & 56.96          & 55.84          & 67.92      \\ \hline
LR = 0.4, MBS = 64                                                       & 58.96          & 65.34          & 68.44      \\ \hline
LR = 0.4, MBS = 256                                                      & 54.20          & 60.50          & 67.80      \\ \hline
\end{tabular}
}
% \vspace{-.15in}
\end{table}

% \vspace{-.1in}
\begin{table}[t]
\caption{{Experimental results from diverse hyperparameters utilized in linear evaluation. Text highlighted in \textbf{bold} indicates the results obtained using default hyperparameters.}}
% \vspace{-.15in}
\label{table:hyperparameter2}
\centering
\smallskip\noindent
% \vspace{-.1in}
\resizebox{.5\linewidth}{!}{
\begin{tabular}{|c||c|c|c|}
\hline
\begin{tabular}[c]{@{}c@{}}Hyperparameter set \\ (for linear eval.)\end{tabular} & FT             & MoCo + CaSSLe  & MoCo + PNR     \\ \hline \hline
LR = 3.0, MBS = 64                                                               & 57.34          & 64.02          & 68.16          \\ \hline
\textbf{LR = 3.0, MBS = 128}                                                     & \textbf{57.72} & \textbf{63.88} & \textbf{68.01} \\ \hline
LR = 3.0, MBS = 256                                                              & 57.52          & 62.96          & 67.78          \\ \hline
LR = 3.0, MBS = 512                                                              & 56.92          & 62.34          & 63.08          \\ \hline
LR = 1.0, MBS = 256                                                              & 56.42          & 62.00          & 66.52          \\ \hline
LR = 7.0, MBS = 256                                                              & 57.02          & 63.88          & 67.98          \\ \hline
\end{tabular}
}
% \vspace{-.15in}
\end{table}

\subsection{Additional Experimental Analysis for Barlow/SimCLR + CaSSLe}\label{sec:analysis}

% (c) $(S, P)=(0.022, 0.018)
 {Figure \ref{figure:each_task_graph3} presents an additional experimental analysis focusing on "Barlow + CaSSLe" and "SimCLR + CaSSLe" in Class-IL (5T) experiments with the ImageNet-100 dataset. Consistent with our previous observations, we once again note an improvement in their Avg(a$_{1:5,t}$); however, certain tasks exhibit signs of catastrophic forgetting. Specifically, "Barlow + CaSSLe" demonstrates a decline in performance for a$_{k=1,t}$ and a$_{k=3,t}$. Similarly, a$_{k=1,t}$ and a$_{k=4,t}$ of "SimCLR + CaSSLe" follow a comparable trend.}

\begin{figure*}[t]
\vspace{-.1in}
\centering 
\subfigure[Barlow + CaSSLe]
{\includegraphics[width=0.32\linewidth]{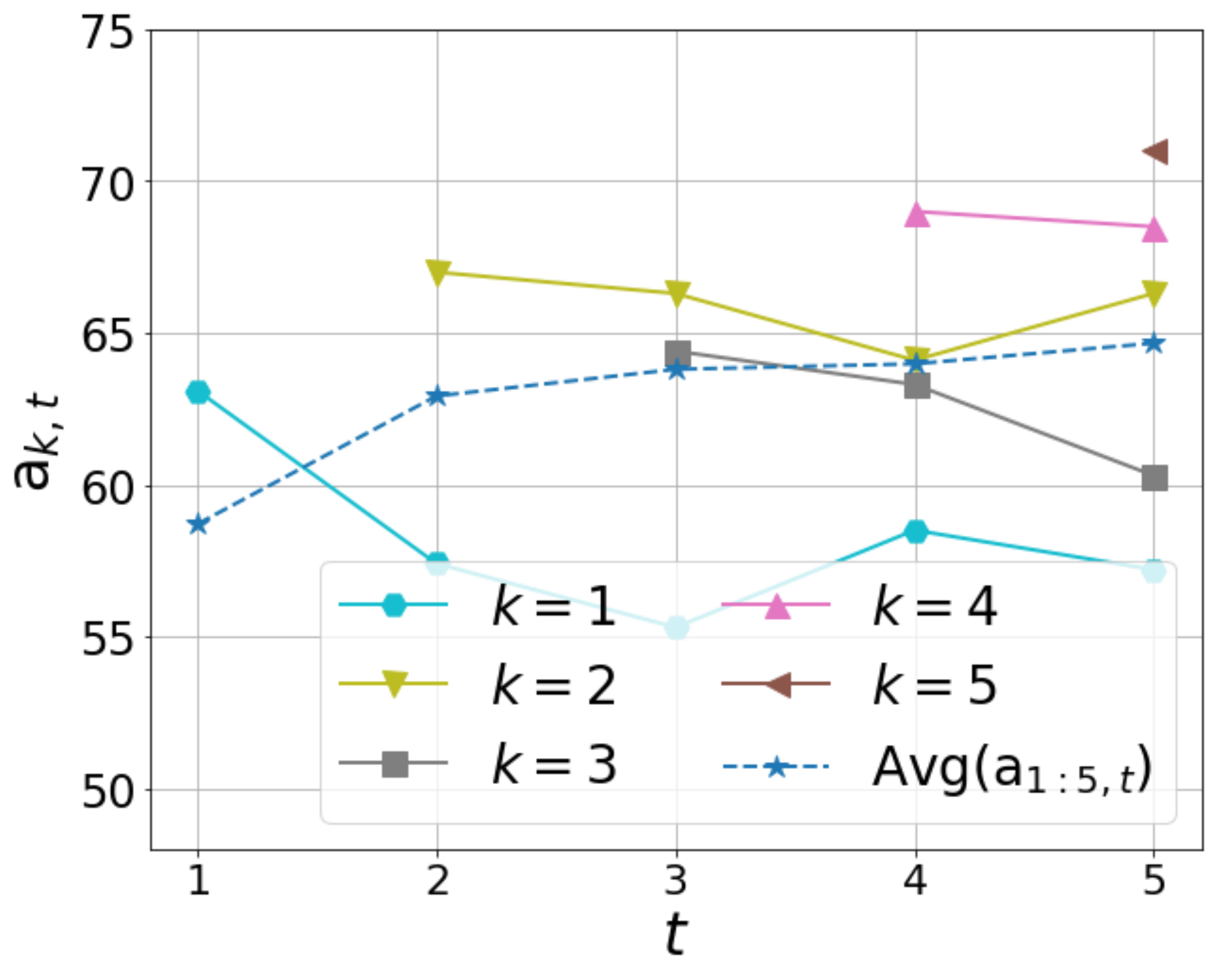}\label{figure:recon_exp_fig4_3}}
\subfigure[SimCLR + CaSSLe]
{\includegraphics[width=0.32\linewidth]{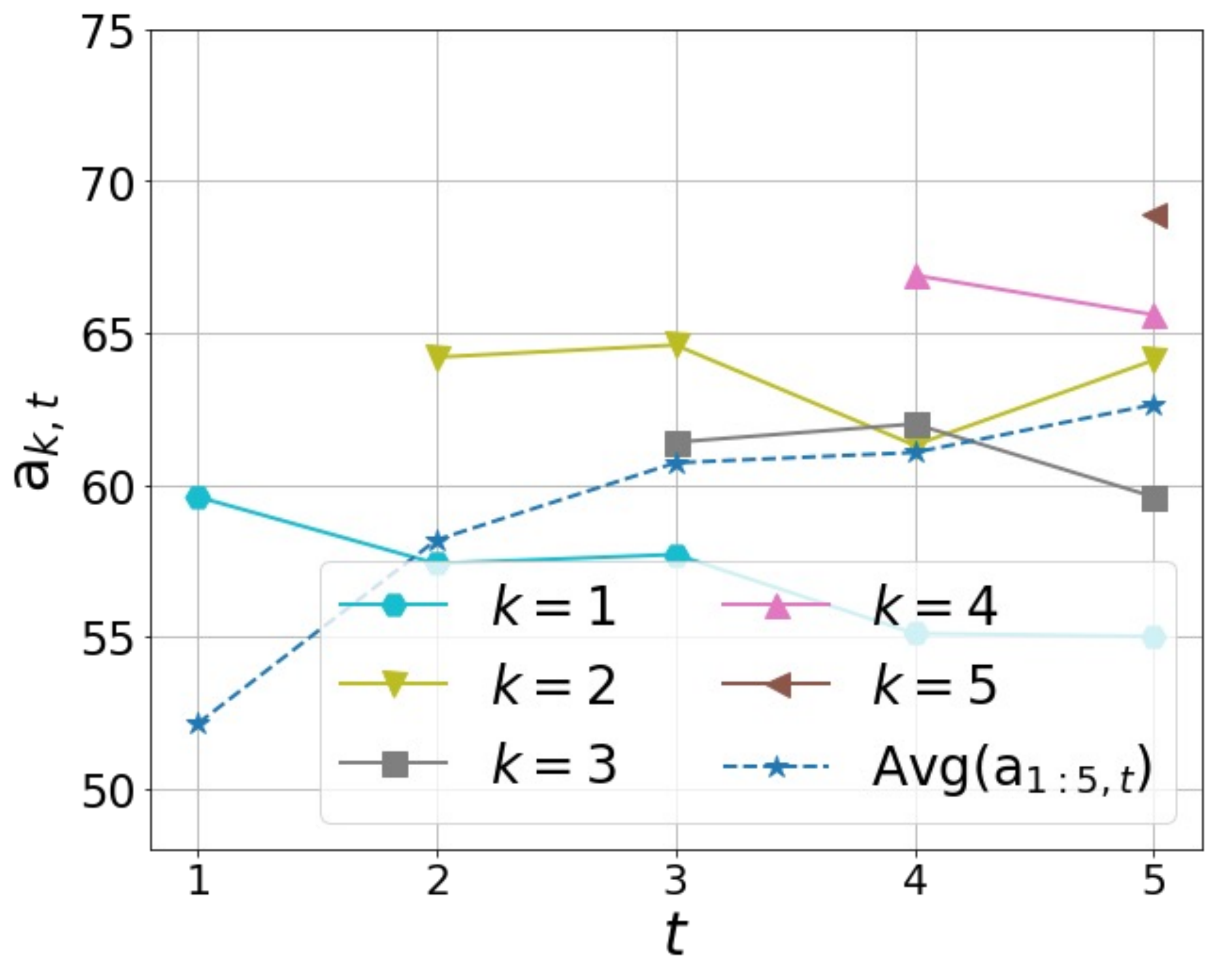}\label{figure:recon_exp_fig4_6}}
% \subfigure[\ {BYOL + FT}]
% {\includegraphics[width=0.32\linewidth]{figures/recon_exp_fig4_7.pdf}\label{figure:recon_exp_fig4_6}}
\vspace{-.1in}
\caption{The graph illustrates the values of a$_{k,t}$ for each algorithm in the Class-IL (5T) scenario. The measured stability and plasticity for each method are as follows: (a) $(S, P)=(2.52, 2.8)$, (b) $(S, P)=(2.22, 1.95)$.}
\label{figure:each_task_graph3}
\vspace{-.1in}
 \end{figure*}

% \newpage

\subsection{{Additional Experiments Using Supervised Contrastive Learning}}\label{sec:supcon}

We introduce Supervised Contrastive Learning (SupCon)~\cite{(sup_contrastive)khosla2020supervised} as an additional method to investigate the effectiveness of PNR in scenarios where labels are available. To accomplish this, we modify the implementation of "SimCLR + PNR" in Equation (\ref{eqn:plasticity}) and (\ref{eqn:stability}) adapting it to utilize supervised labels. More specifically, it incorporates additional negatives in "SupCon + CaSSLe".

% For the Non-contrastive learning-based algorithm, we selected the recent VICReg~\cite{(vicreg)bardes2022vicreg} algorithm to explore the potential application of PNR to various SSL algorithms. To implement PNR, we added additional regularization proposed in Equation (\ref{eqn:PNR_vicreg}) and conducted experiments accordingly.

We conduct experiments with the CIFAR-100 and ImageNet-100 datasets, employing the Class-IL (5T, 10T) scenario as our experimental setting, and the results are shown in Table \ref{table:downstream3}.

\begin{table}[h]
% \vspace{-.1in}
\caption{Experimental results with Supervised Contrastive Learning.}
% \vspace{-.1in}
\label{table:downstream3}
\centering
\smallskip\noindent
\resizebox{.6\linewidth}{!}{
\begin{tabular}{|cc||cc|cc|}
\hline
\multicolumn{2}{|c||}{\multirow{2}{*}{A$_{T}$}}                  & \multicolumn{2}{c|}{CIFAR-100}                       & \multicolumn{2}{c|}{ImageNet-100}                    \\ \cline{3-6} 
\multicolumn{2}{|c||}{}                                          & \multicolumn{1}{c|}{Class-IL 5T}    & Class-IL 10T   & \multicolumn{1}{c|}{Class-IL 5T}    & Class-IL 10T   \\ \hline \hline
\multicolumn{1}{|c|}{\multirow{2}{*}{SupCon}} & CaSSLe          & \multicolumn{1}{c|}{60.38}          & 55.38          & \multicolumn{1}{c|}{66.26}          & 60.48          \\ \cline{2-6} 
\multicolumn{1}{|c|}{}                        & \textbf{PNR} & \multicolumn{1}{c|}{\textbf{60.73}} & \textbf{56.16} & \multicolumn{1}{c|}{\textbf{66.96}} & \textbf{60.95} \\ \hline
% \multicolumn{1}{|c|}{\multirow{2}{*}{VICReg}} & CaSSLe          & \multicolumn{1}{c|}{53.17}          & 47.54          & \multicolumn{1}{c|}{59.18}          & 49.00          \\ \cline{2-6} 
% \multicolumn{1}{|c|}{}                        & \textbf{PNR} & \multicolumn{1}{c|}{\textbf{55.40}} & \textbf{50.78} & \multicolumn{1}{c|}{\textbf{61.88}} & \textbf{51.82} \\ \hline
\end{tabular}
}
% \vspace{-.1in}
\end{table}

Based on the results in the table above, we observe that PNR can be applied to supervised contrastive learning, leading to consistent performance improvements compared to CaSSLe. We believe these experimental findings suggest the potential for the proposed PNR concept to be widely applicable across various domains using contrastive learning-based loss functions in both supervised and self-supervised manners.

\subsection{Detailed Experimental Results of Downstream Tasks}\label{sec:downstream}

\begin{table}[h]
% \vspace{-.1in}
\caption{Experimental results of three downstream tasks.}
% \vspace{-.1in}
\label{table:downstream1}
\centering
\smallskip\noindent
\resizebox{.8\linewidth}{!}{
\begin{tabular}{|cc||c|c|c|c|c||c|c|}
\hline
\multicolumn{2}{|c||}{Scenario}                                         & Downstream       & \begin{tabular}[c]{@{}c@{}}MoCo\\  + CaSSLe\end{tabular} & \begin{tabular}[c]{@{}c@{}}SimCLR\\  + CaSSLe\end{tabular} & \begin{tabular}[c]{@{}c@{}}Barlow\\  + CaSSLe\end{tabular} & \begin{tabular}[c]{@{}c@{}}BYOL\\  + CaSSLe\end{tabular} & \begin{tabular}[c]{@{}c@{}}MoCo\\+ PNR\end{tabular} & \begin{tabular}[c]{@{}c@{}}BYOL\\+ PNR\end{tabular} \\ \hline \hline
\multicolumn{1}{|c|}{\multirow{8}{*}{Class-IL}} & \multirow{4}{*}{5T}  & CIFAR-100        & 44.52                                                    & 41.99                                                      & 43.59                                                      & 46.71                                                    & 47.8                                                   & 48.8                                                   \\ \cline{3-9} 
\multicolumn{1}{|c|}{}                          &                      & CIFAR-10         & 68.09                                                    & 64.9                                                       & 65.43                                                      & 69.42                                                    & 71.26                                                  & 70.21                                                  \\ \cline{3-9} 
\multicolumn{1}{|c|}{}                          &                      & STL-10           & 63.23                                                    & 63.3                                                       & 65.39                                                      & 67.93                                                    & 67.07                                                  & 68.59                                                  \\ \cline{3-9} 
\multicolumn{1}{|c|}{}                          &                      & \textbf{Average} & \textbf{58.61}                                           & \textbf{56.73}                                             & \textbf{58.13}                                             & \textbf{61.35}                                           & \textbf{62.04}                                         & \textbf{62.53}                                         \\ \cline{2-9} 
\multicolumn{1}{|c|}{}                          & \multirow{4}{*}{10T} & CIFAR-100        & 40.84                                                    & 40.34                                                      & 40.67                                                      & 45.49                                                    & 45.86                                                  & 48.63                                                  \\ \cline{3-9} 
\multicolumn{1}{|c|}{}                          &                      & CIFAR-10         & 66.01                                                    & 64.62                                                      & 64.42                                                      & 68.02                                                    & 68.65                                                  & 70.34                                                  \\ \cline{3-9} 
\multicolumn{1}{|c|}{}                          &                      & STL-10           & 60.16                                                    & 59.77                                                      & 62.38                                                      & 65.55                                                    & 65.78                                                  & 67.51                                                  \\ \cline{3-9} 
\multicolumn{1}{|c|}{}                          &                      & \textbf{Average} & \textbf{55.67}                                           & \textbf{54.91}                                             & \textbf{55.82}                                             & \textbf{59.69}                                           & \textbf{60.10}                                         & \textbf{62.16}                                         \\ \hline
\multicolumn{1}{|c|}{\multirow{8}{*}{Data-IL}}  & \multirow{4}{*}{5T}  & CIFAR-100        & 44.5                                                     & 42.33                                                      & 44.5                                                       & 47.02                                                    & 47.07                                                  & 48.14                                                  \\ \cline{3-9} 
\multicolumn{1}{|c|}{}                          &                      & CIFAR-10         & 68.47                                                    & 65.03                                                      & 66.92                                                      & 69.14                                                    & 70.01                                                  & 69.03                                                  \\ \cline{3-9} 
\multicolumn{1}{|c|}{}                          &                      & STL-10           & 64.49                                                    & 63.54                                                      & 67.2                                                       & 68.66                                                    & 69.98                                                  & 68.68                                                  \\ \cline{3-9} 
\multicolumn{1}{|c|}{}                          &                      & \textbf{Average} & \textbf{59.15}                                           & \textbf{56.97}                                             & \textbf{59.54}                                             & \textbf{61.61}                                           & \textbf{62.35}                                         & \textbf{61.95}                                         \\ \cline{2-9} 
\multicolumn{1}{|c|}{}                          & \multirow{4}{*}{10T} & CIFAR-100        & 42.17                                                    & 41.3                                                       & 43.53                                                      & 45.3                                                     & 46.88                                                  & 48.05                                                  \\ \cline{3-9} 
\multicolumn{1}{|c|}{}                          &                      & CIFAR-10         & 66.66                                                    & 65.9                                                       & 65.38                                                      & 69.27                                                    & 69.84                                                  & 70.80                                                  \\ \cline{3-9} 
\multicolumn{1}{|c|}{}                          &                      & STL-10           & 60.35                                                    & 61.88                                                      & 64.51                                                      & 67.35                                                    & 67.48                                                  & 67.95                                                  \\ \cline{3-9} 
\multicolumn{1}{|c|}{}                          &                      & \textbf{Average} & \textbf{56.39}                                           & \textbf{56.36}                                             & \textbf{57.81}                                             & \textbf{60.64}                                           & \textbf{61.40}                                         & \textbf{62.27}                                                  \\ \hline
\end{tabular}
}
% \vspace{-.1in}
\end{table}

As emphasized in several papers~\citep{(repeval)cha2023objective, (simclr)chen2020simple, (moco)he2020momentum}, evaluating the generalization of learned representations across diverse downstream tasks is critical. In line with this, we conduct evaluations on encoders trained with each CSSL scenario on the ImageNet-100 dataset. Following the methodology outlined in \citep{(repeval)cha2023objective}, we use the resized CIFAR-10/-100 datasets (resized to 96x96)~\citep{(cifar)krizhevsky2009learning} and the STL-10 dataset~\citep{(stl10)coates2011analysis} as downstream tasks and perform linear evaluations on them. Table \ref{table:downstream1} showcases the exceptional performance of PNR across various downstream task datasets, consistently achieving the best overall results. Furthermore, the proposed PNR exhibits superior CSSL compared to other CaSSLe variations, particularly evident in the Data-IL scenario.

\begin{table}[h]
% \vspace{-.1in}
\caption{Experimental results of Clipart in DomainNet.}
\label{table:downstream2}
% \vspace{-.1in}
\centering
\smallskip\noindent
\resizebox{.7\linewidth}{!}{
\begin{tabular}{|c||c|c|c|c|c|c|c|}
\hline
Scenario                 &  & \begin{tabular}[c]{@{}c@{}}MoCo\\  + CaSSLe\end{tabular} & \begin{tabular}[c]{@{}c@{}}SimCLR\\  + CaSSLe\end{tabular} & \begin{tabular}[c]{@{}c@{}}Barlow\\  + CaSSLe\end{tabular} & \begin{tabular}[c]{@{}c@{}}BYOL\\  + CaSSLe\end{tabular} & \textbf{\begin{tabular}[c]{@{}c@{}}MoCo\\ + PNR\end{tabular}} & \textbf{\begin{tabular}[c]{@{}c@{}}BYOL\\ + PNR\end{tabular}} \\ \hline \hline
                           & 5T       & 28.32                                                    & 34.68                                                      & 37.42                                                  & 38.98                                                    & {\color[HTML]{333333} \textbf{38.86}}                            & {\color[HTML]{333333} \textbf{41.57}}                            \\ \cline{2-8} 
\multirow{-2}{*}{Class-IL} & 10T      & {\color[HTML]{333333} 28.57}                             & {\color[HTML]{333333} 33.33}                               & {\color[HTML]{333333} 38.30}                           & 38.05                                                    & {\color[HTML]{333333} \textbf{38.19}}                            & {\color[HTML]{333333} \textbf{39.84}}                            \\ \hline
                           & 5T       & {\color[HTML]{333333} 29.74}                             & {\color[HTML]{333333} 34.17}                               & {\color[HTML]{333333} 36.13}                           & 37.04                                                    & {\color[HTML]{333333} \textbf{38.33}}                            & {\color[HTML]{333333} \textbf{40.06}}                            \\ \cline{2-8} 
\multirow{-2}{*}{Data-IL}  & 10T      & {\color[HTML]{333333} 32.81}                             & {\color[HTML]{333333} 35.53}                               & {\color[HTML]{333333} 38.45}                           & 35.91                                                    & {\color[HTML]{333333} \textbf{40.45}}                            & {\color[HTML]{333333} \textbf{37.83}}                            \\ \hline
\end{tabular}
}
\vspace{-.1in}
\end{table}

Following the CaSSLe paper~\citep{(cassle)fini2022self}, 
we conduct additional experiments for downstream tasks. We train a model  with each CSSL scenario on the ImageNet-100 dataset and conduct linear evaluation using the Clipart dataset from DomainNet as the downstream task.
Table \ref{table:downstream2} presents experimental results.
In the scenario of Class-IL, we observe that the models trained with "Barlow + CaSSLe" or "BYOL + CaSSLe" achieve superior performance among baselines. However, "MoCo + PNR" and "BYOL + PNR" also show competitive or state-of-the-art performance, especially in "BYOL + PNR" in Class-IL (5T).
In the case of Data-IL, similar to the results obtained from previous downstream task experiments, we observe that "MoCo + PNR" and "BYOL + PNR" outperform other algorithms by a considerable margin, except for "BYOL + PNR" in Data-IL (10T). 

\subsection{Experimental Results Using a Different Mini-Batch Size}

\begin{table}[h]
\vspace{-.1in}
\caption{Experimental results ($A_{5}$) of Class-IL (5T) with the ImageNet-100 dataset.}
\label{table:batch_size}
\centering
\smallskip\noindent
% \vspace{-.2in}
\resizebox{.85\linewidth}{!}{
\begin{tabular}{|c|c|c|c|c|c|c|c||c|c|}
\hline
\begin{tabular}[c]{@{}c@{}}MoCo\\ + FT\end{tabular} & \begin{tabular}[c]{@{}c@{}}MoCo\\ + CaSSLe\end{tabular} & \begin{tabular}[c]{@{}c@{}}SimCLR\\ + FT\end{tabular} & \begin{tabular}[c]{@{}c@{}}SimCLR\\ + CaSSLe\end{tabular} & \begin{tabular}[c]{@{}c@{}}Barlow\\ + FT\end{tabular} & \begin{tabular}[c]{@{}c@{}}Barlow\\ + CaSSLe\end{tabular} & \begin{tabular}[c]{@{}c@{}}BYOL\\ + FT\end{tabular} & \begin{tabular}[c]{@{}c@{}}BYOL\\ + CaSSLe\end{tabular} & \begin{tabular}[c]{@{}c@{}}MoCo\\ + PNR\end{tabular} & \begin{tabular}[c]{@{}c@{}}BYOL\\+PNR\end{tabular}\\ \hline \hline
55.80                                               & 61.70                                                   & 53.70                                                 & 62.40                                                     & {\color[HTML]{333333} 59.00}                          & {\color[HTML]{333333} 65.00}                              & 59.20                                               & 61.60                                                   & 67.40    & 63.34                                                \\ \hline
\end{tabular}
}
% \vspace{-.1in}
\end{table}

To compare the sensitivity of CaSSLe and PNR to mini-batch size, we train a model with a mini-batch size of 256 (default is 128) and conduct linear evaluation. The experimental results in Table \ref{table:batch_size} show slightly lower results than those mentioned in the manuscript, confirming that increasing the mini-batch size made it challenging to achieve enhanced performance. However, our PNR combination not only demonstrates similar performance across two different mini-batch sizes but also achieves superior performance in both sizes compared to CaSSLe.

% \newpage

\section{Experimental Details}\label{sec:details}

\begin{table}[h]
% \vspace{-.1in}
\caption{Details on training hyperparameters used for CSSL (CIFAR-100 / ImageNet-100 / DomainNet).}
% \vspace{-.1in}
\label{table:details_cssl}
\centering
\smallskip\noindent
\resizebox{.98\linewidth}{!}{
\begin{tabular}{c||ccccc}
\hline
\begin{tabular}[c]{@{}c@{}}CIFAR-100 / ImageNet-100\\  / DomainNet\end{tabular} & \multicolumn{1}{c|}{\begin{tabular}[c]{@{}c@{}}MoCo \\ (+CaSSLe)\end{tabular}} & \multicolumn{1}{c|}{\begin{tabular}[c]{@{}c@{}}SimCLR \\ (+CaSSLe)\end{tabular}} & \multicolumn{1}{c|}{\begin{tabular}[c]{@{}c@{}}BarLow \\ (+CaSSLe)\end{tabular}} & \multicolumn{1}{c|}{\begin{tabular}[c]{@{}c@{}}BYOL\\  (+CaSSLe)\end{tabular}} & {\begin{tabular}[c]{@{}c@{}}All methods\\   + PNR\end{tabular}} \\ \hline \hline
\begin{tabular}[c]{@{}c@{}}Epoch\\  (per task)\end{tabular}                     & \multicolumn{5}{c}{500 / 400 / 200}                                                                                                                                                                                                                                                                                                           \\ \hline
Batch size                                                                      & \multicolumn{5}{c}{256 / 128 / 128}                                                                                                                                                                                                                                                                                                           \\ \hline
Learning rate                                                                   & \multicolumn{1}{c|}{0.4}                                                       & \multicolumn{1}{c|}{0.4}                                                         & \multicolumn{1}{c|}{0.3 / 0.4 / 0.4}                                             & \multicolumn{1}{c|}{1.0 / 0.6 / 0.6}                                           & {\begin{tabular}[c]{@{}c@{}} 0.6 (for MoCo) \\ 0.6 (for SimCLR)\\  0.3 / 0.4 / 0.4 (for Barlow) \\ 1.0 / 0.6 / 0.6 (for BYOL)\end{tabular}}    \\ \hline
Optimizer                                                                       & \multicolumn{5}{c}{SGD}                                                                                                                                                                                                                                                                                                                       \\ \hline
Weight decay                                                                    & \multicolumn{1}{c|}{1e-4}                                                      & \multicolumn{1}{c|}{1e-4}                                                        & \multicolumn{1}{c|}{1e-4}                                                        & \multicolumn{1}{c|}{1e-5}                                                      & {\begin{tabular}[c]{@{}c@{}}1e-5 (for BYOL)\\ 1e-4 (for the others)\end{tabular}}   \\ \hline
\begin{tabular}[c]{@{}c@{}}MLP Layer  (dim)\end{tabular}               & \multicolumn{1}{c|}{2048}                                                      & \multicolumn{1}{c|}{2048}                                                        & \multicolumn{1}{c|}{2048}                                                        & \multicolumn{1}{c|}{4096 / 8192 / 8192}                                                      & {\begin{tabular}[c]{@{}c@{}} 4096 / 8192 / 8192 (for BYOL) \\ 2048 (for the others)\end{tabular}}   \\ \hline
% \begin{tabular}[c]{@{}c@{}}Prediction layer\\  (dim)\end{tabular}     & \multicolumn{5}{c}{2048 / 4096}                                                                                                                                                                                                                                                                                                                      \\ \hline
\begin{tabular}[c]{@{}c@{}}Prediction layer  (dim)\end{tabular}     & \multicolumn{1}{c|}{-}                                                         & \multicolumn{1}{c|}{-}                                                           & \multicolumn{1}{c|}{-}                                                           & \multicolumn{1}{c|}{4096 / 8192 / 8192}                                        & {\begin{tabular}[c]{@{}c@{}} 4096 / 8192 / 8192 (for BYOL) \\ 2048 (for the others)\end{tabular}}      \\ \hline
Queue                                                                           & \multicolumn{1}{c|}{65536}                                                     & \multicolumn{1}{c|}{-}                                                           & \multicolumn{1}{c|}{-}                                                           & \multicolumn{1}{c|}{-}                                                         & {\begin{tabular}[c]{@{}c@{}}65536  (for MoCo)\end{tabular}}  \\ \hline
\begin{tabular}[c]{@{}c@{}}Temperature  ($\tau$)\end{tabular}                 & \multicolumn{1}{c|}{0.2}                                                       & \multicolumn{1}{c|}{0.2}                                                         & \multicolumn{1}{c|}{-}                                                           & \multicolumn{1}{c|}{-}                                                         & {\begin{tabular}[c]{@{}c@{}}0.2  (for MoCo and SimCLR)\end{tabular}}    \\ \hline
\end{tabular}
}
% \vspace{-.1in}
\end{table}

Table \ref{table:details_cssl} presents the training details for each algorithm utilized in our Continual Self-supervised Learning (CSSL) experiments. 
All experiments are conducted on an NVIDIA RTX A5000 with CUDA 11.2 and we follow the experimental settings proposed in the CaSSLe's code~\cite{(cassle)fini2022self}.  
We employ LARS~\citep{(lars)you2017large} to train a model during CSSL. 
% It is worth emphasizing that we strictly followed the training settings of BYOL, as implemented in CaSSLe, including crucial factors such as learning rate, weight decay, learning rate schedule, and augmentations. Notably, we did not perform an extensive search for hyperparameters.

% It is important to mention that all self-supervised learning algorithms use a Projection layer~\cite{(simclr)chen2020simple}, consisting of two layers of MLP with ReLU activation. Specifically, BYOL incorporates a larger default dimension for the Projection layer (dim = 4096) and includes its own Prediction layer (for BYOL) that is not utilized by other self-supervised learning algorithms. This indicates that BYOL assigns a larger portion of weights to the MLP layer in the CSSL process. The subsequent sections will present the consistent results of PNR, accounting for these factors.

% \noindent{\textbf{Domain-IL}} \ \ For the Domain-incremental learning (Domain-IL) experiments conducted on the DomainNet dataset, we follow the experimental setup described in the CaSSLe paper~\cite{(cassle)fini2022self}. Specifically, we perform Domain-IL in the order of Real => QuickDraw => Painting => Sketch => InfoGraph => Clipart. Also, we present the average top-1 accuracy attained by training a linear classifier independently on each domain, utilizing a frozen feature extractor (domain-aware).

% \subsection{Other experimental details}

\noindent{\textbf{$\lambda$ in Equation (\ref{eqn:noncont_l2_2})}} \ \ Table \ref{table:lambda} presents hyperparameter $\lambda$ in Equation (\ref{eqn:noncont_l2_2}) used for experiments.

\begin{table}[h]
\vspace{-.1in}
\caption{Hyperparameter $\lambda$.}
\label{table:lambda}
\centering
\smallskip\noindent
\vspace{-.1in}
\resizebox{.5\linewidth}{!}{
\begin{tabular}{|c||cc|cccc|c|}
\hline
\multirow{3}{*}{$\lambda$} & \multicolumn{2}{c|}{CIFAR-100} & \multicolumn{4}{c|}{ImageNet-100}                                                   & DomainNet \\ \cline{2-8} 
                           & \multicolumn{2}{c|}{Class-IL}  & \multicolumn{2}{c|}{Class-IL}                      & \multicolumn{2}{c|}{Data-IL}   & Domain-IL \\ \cline{2-8} 
                           & \multicolumn{1}{c|}{5T}  & 10T & \multicolumn{1}{c|}{5T} & \multicolumn{1}{c|}{10T} & \multicolumn{1}{c|}{5T}  & 10T & 6T        \\ \hline \hline
Barlow                     & \multicolumn{1}{c|}{1}   & 1   & \multicolumn{1}{c|}{1}  & \multicolumn{1}{c|}{1}   & \multicolumn{1}{c|}{0.1}   & 0.1   & 1         \\ \hline
BYOL                       & \multicolumn{1}{c|}{0.5}   & 0.7   & \multicolumn{1}{c|}{1}  & \multicolumn{1}{c|}{1}   & \multicolumn{1}{c|}{0.1} & 0.1 & 1         \\ \hline
VICReg                     & \multicolumn{1}{c|}{23}  & 23  & \multicolumn{1}{c|}{23}  & \multicolumn{1}{c|}{23}   & \multicolumn{1}{c|}{5}  & 5  & 8         \\ \hline
\end{tabular}
}
% \vspace{-.1in}
\end{table}

\noindent{\textbf{Linear evaluation}} \ \ Table \ref{table:details_linear} presents detailed training hyperparameters employed for linear evaluation on each dataset using encoders trained via CSSL. In the case of CaSSLe variations, we conduct the experiments while maintaining consistent linear evaluation settings with those employed in CaSSLe.

\begin{table}[h]
\vspace{-.1in}
\caption{Experimental details of linear evaluation (CIFAR-100 / ImageNet-100 / DomainNet). }
\label{table:details_linear}
\centering
\smallskip\noindent
\vspace{-.1in}
\resizebox{.8\linewidth}{!}{
\begin{tabular}{c||ccccc}
\hline
\begin{tabular}[c]{@{}c@{}}CIFAR-100 / ImageNet-100\\  / DomainNet\end{tabular} & \multicolumn{1}{c|}{\begin{tabular}[c]{@{}c@{}}MoCo \\ (+CaSSLe)\end{tabular}} & \multicolumn{1}{c|}{\begin{tabular}[c]{@{}c@{}}SimCLR \\ (+CaSSLe)\end{tabular}} & \multicolumn{1}{c|}{\begin{tabular}[c]{@{}c@{}}BarLow \\ (+CaSSLe)\end{tabular}} & \multicolumn{1}{c|}{\begin{tabular}[c]{@{}c@{}}BYOL\\  (+CaSSLe)\end{tabular}} & PNR \\ \hline \hline
\begin{tabular}[c]{@{}c@{}}Epoch\\  (per task)\end{tabular}                     & \multicolumn{5}{c}{100}                                                                                                                                                                                                                                                                                                   \\ \hline
Batch size                                                                      & \multicolumn{5}{c}{128 / 256 / 256}                                                                                                                                                                                                                                                                                                           \\ \hline
Learning rate                                                                     & \multicolumn{1}{c|}{3.0}                                                       & \multicolumn{1}{c|}{1.0}                                                         & \multicolumn{1}{c|}{0.1}                                                         & \multicolumn{1}{c|}{3.0}                                                       & 3.0   \\ \hline
Scheduler                                                                       & \multicolumn{5}{c}{Step LR (steps = {[}60, 80{]}, gamma = 0.1)}                                                                                                                                                                                                                                                                               \\ \hline
Optimizer                                                                       & \multicolumn{5}{c}{SGD}                                                                                                                                                                                                                                                                                                                       \\ \hline
Weight decay                                                                    & \multicolumn{5}{c}{0}                                                                                                                                                                                                                                                                                                                         \\ \hline
\end{tabular}
}
% \vspace{-.1in}
\end{table}

\noindent{\textbf{Downstream task}} \ \ 
Table \ref{table:details_downstream} outlines detailed training hyperparameters employed for linear evaluation on downstream tasks using encoders trained under various CSSL scenarios with the ImageNet-100 dataset. 
% Consistent with CaSSLe-based algorithms, we adhered to the experimental settings detailed in their respective papers. 

\begin{table}[h]
\vspace{-.2in}
\caption{Experimental details of linear evaluation on downstream tasks.}
\label{table:details_downstream}
\centering
\smallskip\noindent
% \vspace{-.2in}
\resizebox{.8\linewidth}{!}{
\begin{tabular}{c||ccccc}
\hline
\begin{tabular}[c]{@{}c@{}}CIFAR-10 / CIFAR-100 / STL-10\\ / Clipart\end{tabular} & \multicolumn{1}{c|}{\begin{tabular}[c]{@{}c@{}}MoCo \\ (+CaSSLe)\end{tabular}} & \multicolumn{1}{c|}{\begin{tabular}[c]{@{}c@{}}SimCLR \\ (+CaSSLe)\end{tabular}} & \multicolumn{1}{c|}{\begin{tabular}[c]{@{}c@{}}BarLow \\ (+CaSSLe)\end{tabular}} & \multicolumn{1}{c|}{\begin{tabular}[c]{@{}c@{}}BYOL\\  (+CaSSLe)\end{tabular}} & PNR \\ \hline \hline
\begin{tabular}[c]{@{}c@{}}Epoch\\  (per task)\end{tabular}                       & \multicolumn{5}{c}{100}                                                                                                                                                                                                                                                                                                                       \\ \hline
Batch size                                                                        & \multicolumn{5}{c}{256}                                                                                                                                                                                                                                                                                                                       \\ \hline
Learning rate                                                                     & \multicolumn{1}{c|}{3.0}                                                       & \multicolumn{1}{c|}{1.0}                                                         & \multicolumn{1}{c|}{0.1}                                                         & \multicolumn{1}{c|}{3.0}                                                       & 3.0   \\ \hline
Scheduler                                                                         & \multicolumn{5}{c}{Step LR (steps = {[}60, 80{]}, gamma = 0.1)}                                                                                                                                                                                                                                                                               \\ \hline
Optimizer                                                                         & \multicolumn{5}{c}{SGD}                                                                                                                                                                                                                                                                                                                       \\ \hline
Weight decay                                                                      & \multicolumn{5}{c}{0}                                                                                                                                                                                                                                                                                                                         \\ \hline
\end{tabular}
}
% \vspace{-.1in}
\end{table}

% \section{Additional Experimental Results}

\newpage

\end{document}